\pdfoutput=1
\documentclass[11pt]{article}
\usepackage[final]{acl}

\usepackage{times}
\usepackage{latexsym}
\usepackage[T1]{fontenc}
\usepackage[utf8]{inputenc}
\usepackage{microtype}
\usepackage{inconsolata}
\usepackage{graphicx}

\usepackage{hyperref}
\usepackage{url}
\usepackage{booktabs}
\usepackage{amsthm}
\usepackage[toc,page,header]{appendix}
\usepackage{minitoc}

\usepackage{graphicx} 
\usepackage{amsmath}
\usepackage{caption}
\usepackage{hyperref}
\usepackage{natbib}
\usepackage{wrapfig}
\usepackage{amssymb}
\usepackage{array}
\usepackage{multirow}
\usepackage{multicol}
\usepackage{adjustbox}
\usepackage{makecell}
\usepackage{bbm}

\usepackage{hhline}
\usepackage{float}
\usepackage{caption}
\usepackage{stfloats}

\newcount\Comments
\usepackage{color}
\definecolor{darkgrey}{gray}{0.5}
\definecolor{Green}{rgb}{0.13, 0.55, 0.13}
\definecolor{BrickRed}{rgb}{0.8, 0.25, 0.33}

\newcommand{\kibitz}[2]{\ifnum\Comments=1{\color{#1}{#2}}\fi}
\newcommand{\RR}[1]{\kibitz{orange}{[Roi: #1]}}
\newcommand{\NC}[1]{\kibitz{purple}{[Nitay: #1]}}
\newcommand{\ON}[1]{\kibitz{cyan}{[Omer: #1]}}
\newcommand{\OK}[1]{\kibitz{olive}{[Orgad: #1]}}
\newcommand{\IS}[1]{\kibitz{teal}{[Idan: #1]}}
\newcommand\todo[1]{\kibitz{red}{TODO: {#1}}}

\newcommand\tocitef[1]{\kibitz{red}{[CITE: {#1}]}}

\newcommand\draft[1]{{{#1}}}
\newcommand\greyfont[1]{\kibitz{darkgrey}{{#1}}}

\newcommand{\paragraphs}[1]{\textbf{#1}\hspace{1mm}}

\newcommand{\paragraphn}[1]{\medskip\noindent\textbf{#1}\indent}

\newcommand\codefont[1]{{\fontfamily{qcr}\selectfont
#1}}

\newenvironment{symbolfootnotes}
  {\par\edef\savedfootnotenumber{\number\value{footnote}}
   
   \setcounter{footnote}{0}}
  {\par\setcounter{footnote}{\savedfootnotenumber}}

\title{Are LLMs Better than Reported? \\Detecting Label Errors and Mitigating Their Effect on Model Performance}
\author{\textbf{Omer Nahum}\textsuperscript{\emph{T}} \hspace{5pt}
\textbf{Nitay Calderon}\textsuperscript{\emph{T}} \hspace{5pt}
\textbf{Orgad Keller}\textsuperscript{\emph{G}} \hspace{5pt} 
\textbf{Idan Szpektor}\textsuperscript{\emph{G}}
\hspace{5pt} 
\textbf{Roi Reichart}\textsuperscript{\emph{T}} \hspace{5pt}
\\
\textsuperscript{\emph{T}}Technion - Institute of Technology \hspace{5pt}
\textsuperscript{\emph{G}}Google Research
}

\begin{document}
\maketitle

\doparttoc 
\faketableofcontents 

\begin{abstract}
NLP benchmarks rely on standardized datasets for training and evaluating models and are crucial for advancing the field. Traditionally, expert annotations ensure high-quality labels; however, the cost of expert annotation does not scale well with the growing demand for larger datasets required by modern models.
While crowd-sourcing provides a more scalable solution, it often comes at the expense of annotation precision and consistency. Recent advancements in large language models (LLMs) offer new opportunities to enhance the annotation process, particularly for detecting label errors in existing datasets. In this work, we consider the recent approach of LLM-as-a-judge, leveraging an ensemble of LLMs to flag potentially mislabeled examples.
\draft{We conduct a case study on four factual consistency datasets from the TRUE benchmark, spanning diverse NLP tasks, and on SummEval, which uses Likert-scale ratings of summary quality across multiple dimensions.}
We empirically analyze the labeling quality of existing datasets and compare expert, crowd-sourced, and LLM-based annotations in terms of the agreement, label quality, and efficiency, demonstrating the strengths and limitations of each annotation method. Our findings reveal a substantial number of label errors, which, when corrected, induce a significant upward shift in reported model performance. This suggests that many of the LLMs' so-called mistakes are due to label errors rather than genuine model failures. Additionally, we discuss the implications of mislabeled data and propose methods to mitigate them in training to improve performance. 

\end{abstract}
\section{Introduction}

\begin{figure*}[t]
    \centering
    \includegraphics[width=0.90\linewidth]{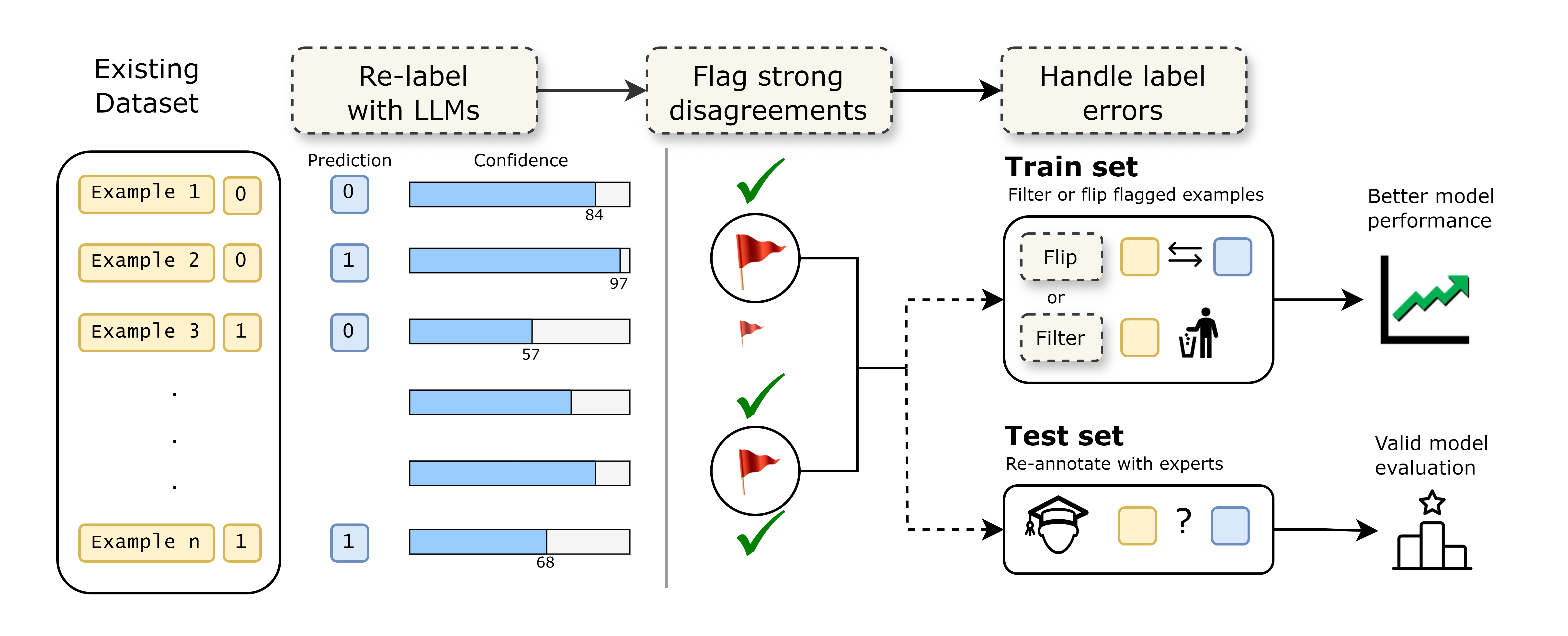}
    \vspace{-0.5em}
    \caption{An illustration of our approach for detecting and addressing mislabeled data: (1) Re-label examples from existing datasets using an ensemble of LLMs. (2) Identify \textit{strong disagreements} between the LLM's predictions and the original labels (i.e., high confidence in a different label), flagging examples based on confidence levels. Our findings show that LLMs detect between 6\% and 21\% of label errors, and higher LLM confidence is strongly associated with improved precision in error detection. (3) In the training set, we either filter or flip flagged examples, leading to an increase of up to 4\%. For the test set, flagged examples are re-annotated by experts to make sure the evaluation is accurate. Under accurate evaluation, the performance of LLMs is up to 15\% higher.}
    \label{fig:flow}
    \vspace{-1em}
\end{figure*}

Natural Language Processing (NLP) benchmarks have long served as a cornerstone for advancing the field, providing standardized datasets for training and evaluating methods and models \citep{GLUE, MMLU, BIG-bench, calderon2024measuring}.
These datasets have been developed over the years for various tasks and scales, annotated using different schemes. Gold labels represent the ``true'' or ground truth annotations, which are typically established through expensive rigorous processes, including expert consensus and extensive quality control. However, as models have increased in size \citep{BERT, GPT3}, the demand for larger datasets has also grown \citep{DBLP:journals/corr/abs-2001-08361}. Since expert annotation is cost-prohibitive, it does not scale well to meet these demands. The demand for large quantities of annotated data quickly and cost-effectively has led researchers to adopt crowd-sourcing, often sacrificing expertise for scale.

That way or another, constructing datasets heavily involves making compromises in annotation, trading off between scale, efficiency and expertise. 
Even when annotated by experts, datasets can naturally contain labeling errors, arising from factors such as task subjectivity, annotator fatigue, inattention, insufficient guidelines, and more \citep{RogersSK13, ReissXCME20, SylolypavanSWS23}. Mislabeled data is even more pronounced when non-expert annotators are involved \citep{kennedy2020shape, ChongHM22}. 
Widespread mislabeled data is particularly concerning because both the research community and the industry rely heavily on benchmarks. In training data, label errors harm model quality and hinder generalization, while in test sets, they lead to flawed comparisons, false conclusions, and prevent progress. 

Recent advancements in LLMs \citep{DBLP:conf/nips/Ouyang0JAWMZASR22, chiang-lee-2023-large, Li2023CoAnnotatingUW, GatCFCSR24} present new opportunities to improve the annotation process, specifically in \draft{detecting label errors within existing datasets \citep{10.1162/coli_a_00464}}. Rather than re-annotating entire datasets (e.g., through experts or crowd-workers), we consider the LLM-as-a-judge approach \citep{llm_as_a_judge}, and propose a simple yet effective method by leveraging an ensemble of LLMs to flag a set of potentially mislabeled examples. These can then be sent to experts for re-annotation and correction, or even get filtered during training.

Specifically, we construct an ensemble model using multiple LLMs with diverse prompts, gathering both their predicted labels and corresponding confidence scores. These predictions are contrasted with the original labels, and instances where the LLMs \emph{strongly disagree} with the original label (i.e., show high confidence in a different label) are flagged as potential mislabeling cases.
Additionally, we not only explore the role of LLMs in detecting errors but also evaluate their performance as annotators, comparing them with expert and crowd-sourced annotations. We assess these approaches in terms of agreement, label quality, and efficiency, highlighting their strengths and limitations.

\draft{To address the broader issue of label errors in NLP benchmarks, we conduct a comprehensive end-to-end study structured around four interconnected research questions:}
(1) Do current benchmarks include mislabeled data? (2) Can LLMs detect label errors? (3) How do expert, crowd-sourced, and LLM-based annotations compare in quality and efficiency? and (4) What are the implications of mislabeled data on model performance and can we mitigate their impact? 

To this end, we choose the TRUE benchmark \citep{TRUE} -- A collection consolidating 11 existing datasets annotated for factual consistency in
a unified format -- as a case-study and empirically investigate its labeling quality. Specifically, we analyze four datasets from TRUE with binary factual consistency annotation originating from different tasks. To support our claims and results in other setups, we conduct similar experiments on an additional dataset, SummEval \citep{summeval}, which evaluates generated summaries in four dimensions on a scale of 1 to 5.

Our paper presents both methodological and empirical contributions. We propose a straightforward approach for detecting potential mislabeled examples (as illustrated in \autoref{fig:flow}), revealing a substantial number of label errors in existing datasets, ranging from 6\% to 21\%. Additionally, we demonstrate that the precision of LLMs in identifying errors improves with their confidence in an incorrect label; when their confidence exceeds 95\%, over two-thirds of those labels are human errors. Moreover, we show that LLM-based annotations not only excel in error detection but also perform similarly to, or better than, traditional annotation methods, offering better trade-offs between quality, scale, and efficiency. Finally, we empirically illustrate the negative impact of mislabeled data on model training and evaluation. We propose a simple automated method for addressing label errors, improving the performance of fine-tuned models by up to 4\%. In evaluation, we found that mislabeled data can significantly distort reported performance; LLMs may perform up to 15\% better. This indicates that many so-called prediction errors are not genuine errors but are instead human annotation mistakes.

\draft{Together, our results offer a holistic perspective on label errors, examining their prevalence in real datasets, the trade-offs and practices that give rise to them, the role LLMs can play across the annotation process, and their downstream effects on model performance.}

\section{Related Work}

\paragraph{Traditional Human Annotation Approaches}
Crowdsourcing is widely used for annotating large-scale NLP datasets \citep{rajpurkar-etal-2016-squad, MNLI, Super-NaturalInstructions}, offering rapid and scalable data collection. However, quality control remains a challenge, with labeling inconsistencies increasing as dataset complexity grows \citep{Quality-Control-of-Crowdsourcing, Quality-Control-in-Crowdsourcing-issues}. Moreover, as LLMs approach near-human performance \citep{chiang-lee-2023-large, chen-ding-2023-probing}, crowd workers increasingly rely on these models for assistance, further complicating annotation quality \citep{artificial_users, veselovsky2023prevalencepreventionlargelanguage}.
Expert annotation provides more reliable labels for domain-specific and cognitively demanding tasks (e.g., medical or legal domains) but is significantly slower and costlier than crowdsourcing \citep{Snow2008CheapAF, Chau2020UnderstandingTT}. Ensuring inter-annotator agreement among experts adds further complexity and expense \citep{Baledent2022ValidityAC}. 
Our study compares expert, crowd-sourced, and LLM-based annotation approaches in terms of quality and efficiency.

\paragraphn{LLMs in the Annotation Loop}
LLMs have been increasingly utilized as annotators in various NLP tasks, offering potential benefits in efficiency and scalability, often outperforming human annotators \citep{He2023AnnoLLMML, Gilardi2023ChatGPTOC, Trnberg2023ChatGPT4OE, calderon2024behalf}.
However, LLMs are not reliable as standalone annotators as they may produce incorrect labels, particularly in complex \citep{Chen2024IsAL}, social \citep{Ventura2023navigating, Felkner2024GPTIN}, emotional \citep{LissakCSOFKR24}, or low-resource \citep{bhat-varma-2023-large} contexts. To mitigate these limitations, hybrid approaches combining LLMs with human oversight have been proposed \citep{Kim2024MEGAnnoAH, Li2023CoAnnotatingUW, weber-plank-2023-activeaed, Zhang2023LLMaAAML, Kholodna2024LLMsIT}.
While most research focuses on annotation from scratch, our work employs an ensemble of LLMs to flag potentially mislabeled data points in existing datasets. \citet{bavaresco-etal-2025-llms} compare LLM- and human-provided annotations, focusing on agreement rather than detecting label errors or analyzing their implications.

\paragraphn{Handling Label Errors} Label errors (also referred to as label noise) in training and evaluation datasets can significantly impair NLP model performance and reliability \citep{Frnay2014ClassificationIT}. Previous work mainly focuses on fine-tuned models and typically identifies mislabeled examples based on the model's low confidence or high training loss \citep{ChongHM22, Hao2020InaccurateLI, Pleiss2020IdentifyingMD, Northcutt2019ConfidentLE}. For example, \citet{ChongHM22} detects label errors using the loss of a fine-tuned model, primarily in binary classification, with some ensemble-based variants explored. Once these high-loss or low-confidence examples are flagged, they are typically filtered out \citep{Nguyen2019SELFLT, Northcutt2019ConfidentLE}, corrected automatically \citep{Pleiss2020IdentifyingMD, Hao2020InaccurateLI}, or re-labeled by human annotators \citep{Northcutt2021PervasiveLE} to verify and improve dataset quality.
Our work differs both methodologically and in scope. We use zero-shot LLMs with prompt diversity to construct an ensemble, requiring no model training, enabling broader adaptability. While prior approaches often flag uncertain predictions, we focus on confident disagreements, where the model strongly favors a different label. This makes the flagged cases more actionable, as they highlight what the model believes the label should be.
Recent work on AED also includes more nuanced views: distinguishing genuine errors from legitimate variation \citep{weber-genzel-etal-2024-varierr}, introducing model-agnostic frameworks that detect and overwrite erroneous labels \citep{yang2023improvingopinionbasedquestionanswering}, and benchmarking AED across tasks and datasets to support reproducibility \citep{10.1162/coli_a_00464}. 

\section{LLM as an Annotator and Detector}
\label{sec:llm_method}
This study aims to evaluate the potential of LLMs in detecting mislabeled examples and compare three annotation approaches: experts, crowdsourcing, and LLMs. To this end, we use an ensemble model that combines multiple LLMs with varied prompts. The motivation for this ensemble is twofold: first, we demonstrate that it enhances error detection and aligns more closely with expert annotations while also decreases the variance; second, it offers a simple approach that avoids the need for complex model selection or extensive prompt engineering, relying instead on the collective strength.

\paragraphn{Prediction and Confidence} To make a prediction using the ensemble, we first extract class probabilities of each LLM and prompt from the logits of the representing class tokens (e.g., 0 or 1 for the binary TRUE datasets, and 1 to 5 for the ordinal SummEval). The probabilities are then normalized to sum to 1. Next, we compute the average probability for each class across the ensemble and select the class with the highest probability (argmax) as the final prediction. 
The confidence in the prediction is defined as the corresponding ensemble probability. If the token probabilities are not accessible, they can be approximated via sampling.

\paragraphn{Errors Detection} We re-label the dataset using the ensemble, keeping both the prediction and confidence for each example. We then flag potentially mislabeled examples where there is \emph{strong disagreement} between the ensemble prediction and the original label, specifically when the model exhibits high confidence in a false prediction. In the binary case, we examine only examples where the ensemble prediction differs from the original label. In the ordinal case, we examine examples where the difference between the original label and the ensemble prediction is strictly greater than 1 (e.g., 3 vs. 5, 1 vs. 5, 4 vs. 2, etc.). After examining these examples, only those with confidence exceeding a predefined threshold are flagged as potentially mislabeled. Our experiments show that as confidence in an incorrect prediction increases, the likelihood of the example being mislabeled also rises.


For test sets, flagged examples can be re-examined by experts to verify their labels. For training sets, the same applies, though automated alternatives can be to remove or relabel them based on the ensemble prediction.
\section{Experimental Setup}
\label{sec: experimental}

\subsection{Data}
As a case-study, we choose to explore the extensive and widely used TRUE benchmark \citep{TRUE}, which is typically used as an evaluation set \citep{steen-etal-2023-little, gekhman-etal-2023-trueteacher, wang-etal-2024-less, Zha2023AlignScoreEF}. It consists of 11 datasets from various NLP tasks such as summarization and knowledge-grounded dialogue. This benchmark is unique in its approach of bringing multiple datasets and tasks into a unified schema of binary factual consistency labels. Each dataset is transformed from its original structure (e.g., a source document and a summary) into two input texts, \emph{Grounding} and \emph{Generated Text}, and a binary label indicating whether the generated text is factually consistent w.r.t the grounding. This enables us to examine multiple tasks and domains under the same umbrella at once while maintaining a unified binary-label schema. 
Specifically, we focus on four TRUE datasets, one from each task: MNBM -- summarization evaluation \citep{MNBM}; BEGIN -- grounded dialogue evaluation \citep{BEGIN}; VitaminC -- fact verification \citep{vitc}; and PAWS -- paraphrasing evaluation \citep{PAWS}. See \autoref{sec: appendix_data} for additional details on these datasets.

For each dataset, we randomly sampled up to 1000 examples (using the full dataset if smaller) for LLM annotation. From these, 160 examples per dataset (640 in total) form the evaluation set, while the remainder were kept for training and validation (\autoref{sec: implications-training}). The evaluation set was further re-annotated by two experts and three crowd workers.

\paragraphn{SummEval} \indent
In addition to the TRUE benchmark, we replicate some of the experiments on the full SummEval benchmark \citep{summeval}. This benchmark includes 1600 generated summaries evaluated on \textit{four dimensions} (relevance, fluency, coherence, consistency) by crowd-workers and experts. 
In contrast to TRUE, the labeling scheme is \textit{ordinal} on a scale of 1 to 5. For further information on the SummEval data and experimental setting, see \autoref{sec: appendix_summeval}. Noteworthy, when researchers employ the SummEval benchmark, they use solely the expert annotations. Accordingly, the focus of our experiments conducted on SummEval is (1) to simulate a setup where the original labels are obtained through crowd-sourcing while relying on expert annotations as the gold standard; and (2) to compare the three annotation approaches (crowd-sourcing, experts, and LLMs).

\subsection{Annotation Procedure}
This subsection outlines the annotation procedures for the various approaches. Refer to \autoref{sec: appendix-annotation} for additional implementation and technical details not covered here, or \autoref{sec: appendix_summeval} for the SummEval LLM annotation details.

\paragraphn{LLMs}
\label{sec: annotation-llm}
We re-annotate the data with four LLMs: GPT-4, \citep{gpt4}, PaLM2 \citep{palm2}, Mistral (7B) \citep{mistral7b}, Llama 3 (8B) \citep{llama3}, and GPT-4o and Gemini-1.5-Flash for SummEval. Our ensemble model leverages four different prompts which control the variance caused by task descriptions. The prompts are designed as a zero-shot classification task, e.g., for TRUE the requested output is a single token, either \codefont{'0'} for factual inconsistency or \codefont{'1'} for factual consistency (see more details in Appendix, \ref{sec: appendix-llms} and prompt templates in \autoref{fig:prompts}).

\paragraphn{Crowd-sourcing}
\label{sec: annotation-crowd-sourcing}
Generally, crowd-sourced annotators span a spectrum-- from untrained, "common" crowd-workers to carefully selected and trained annotators. Our paper focuses on the lower end of this spectrum.
We used Amazon Mechanical Turk (MTurk) to recruit crowd workers for annotating 100 examples per TRUE dataset (400 total). Examples were randomly assigned to annotators.  
Each annotated example was manually reviewed. Rejected examples were returned to the pool and re-annotated until each example was annotated by three different annotators. 

To obtain a single label per example, we consider two different aggregations: (1) \emph{Majority} - by majority vote, and (2) \emph{Strict} - if any annotator marks it \textit{inconsistent}, that becomes the label. For SummEval, we use the crowd-sourced annotations provided by \citet{summeval}, aggregated by their median. 

\paragraph{Experts}
\label{sec: annotation-expert}
All TRUE examples where the prediction differed from the original label, regardless of confidence, were annotated by human experts. The experts are two of the paper's authors, who are familiar with the guidelines and task characteristics.

Each example was independently annotated by both experts on a scale from 0 (\textit{inconsistent}) to 1 (\textit{consistent}). The examples were shuffled and presented in no specific order, with neither the original nor LLM labels shown. For cases where the experts disagreed, a reconciliation phase followed, during which they discussed and attempted to resolve their differences.
For more details on the annotation procedure, see Appendix \ref{appendix-experts}. 
After re-annotating all conflicted examples, we define the \textit{gold label} as the original label, if the LLM prediction agrees with it, or the expert resolution, if there was a disagreement. For SummEval, we use the expert annotations provided by \citet{summeval}, aggregated by median.

\section{Label Errors: Analysis and Detection}
\subsection{Do current benchmarks include mislabeled data?}
\label{sec: label_errors}

To address the first research question, we annotate the test-set of TRUE (as described in \autoref{sec: experimental} using LLMs. We then contrast these annotations with the original labels, to find disagreements. As shown in \autoref{table: llm_dis_err}, the disagreement rate is significant and can be up to $\sim40\%$ of the examples. 
An example of such disagreement is presented in \autoref{table: annotation errors}.
While this would typically suggest poor LLM performance, we further investigated by re-annotating with experts to determine which was more accurate: the original label or the LLMs' prediction.

Our findings show a considerable number of label errors for all examined datasets (see the \%error column in \autoref{table: llm_dis_err}).
Based on the experts \textit{gold label} and the sample sizes, we also estimate a lower bound for the total percentage of label errors in the full datasets. We employed the Clopper-Pearson exact method \citep{clopper-pearson} to construct a 95\% confidence interval for the binomial proportion, adjusted by a finite population correction (FPC) (see more details in Appendix \ref{sec: appendix-CI_error_rate}). We provide the lower bound of these confidence intervals in parentheses in \autoref{table: llm_dis_err}, under the \%error column. The lower bounds range from 3\% in the PAWS dataset to 15.8\% in the BEGIN dataset.

\begin{table}[t]
\centering
\renewcommand{\arraystretch}{1.2}
\begin{adjustbox}{width=\linewidth}
\small 
\begin{tabular}{|p{7cm}|}
    \hline
    \textbf{Dataset:} \small BEGIN \\
    \textbf{Grounding:} \small Hillary Clinton, the nominee of the Democratic Party for president of the United States in 2016, has taken positions on political issues while serving as First Lady of Arkansas (1979–81; 1983–92), First Lady of the United States (1993–2001); 
    \vspace{0.5em}
    \textbf{Generated Text:} \small She is the nominee in 2016. 
    \vspace{0.5em}
    \makecell[l]{\textbf{Original Label:} 0 \hspace{1em} \textbf{LLM $p$:} 0.98 \hspace{1em} \textbf{Gold Label:} 1} \newline
    \greyfont{\textbf{Explanation}: \small She (Hillary Clinton) is indeed the nominee in 2016 as specifically stated in the grounding.} \\
    \hline
\end{tabular}
\end{adjustbox}
\caption{Example of an annotation error in the original datasets, discovered by LLMs and corrected by experts. In Appendix \autoref{table: annotation_errors_full} we provide additional examples.} 
\label{table: annotation errors}
\end{table}

\begin{table}[t]
\centering
\begin{adjustbox}{width=\linewidth}
\begin{tabular}{p{1.5cm}|p{2.5cm}|p{1cm}|p{1.5cm}|p{1.7cm}}
\toprule
\textbf{Dataset} & \textbf{Task} & \textbf{\% pos} & \textbf{\% LLM disagree} & \textbf{\% error} \\
\midrule
MNBM & Summarization & 10.6 & 39.4 & 16.9 \small{(11.6)}\\
BEGIN & Dialogue & 38.7 & 34.4 &  21.2 \small{(15.8)} \\
VitaminC & Fact Verification & 52.5 & 17.5 & 8.1 \small{(4.4)}\\
PAWS & Paraphrasing & 44.3 & 22.5 & 6.2 \small{(3.0)}\\
\bottomrule
\end{tabular}
\end{adjustbox}
\caption{Summary of LLM disagreement and label error rates across different datasets. \%pos is the percentage of positive (i.e., the \textit{consistent} class) examples in the data. \% LLM disagree refers to the percentage of examples where the LLM label differs from the original one. \% error indicates the error rate in the sampled test set, while the number in parentheses denotes the estimated lower bound of the error rate for the entire dataset.}
\label{table: llm_dis_err}
\vspace{-1em}
\end{table}


\subsection{Can LLMs Detect Label Errors?}
\label{sec: RQ2}
As described in \autoref{sec: label_errors}, we utilize LLMs to flag candidates for mislabeling, and indeed find label errors. In this subsection, we focus on the LLM viewpoint, exploring the effect of LLM confidence, and the power of ensemble.

\paragraphn{Confidence} LLM annotations are valuable for flagging mislabeled data, offering more than just hard labels. By considering LLM confidence scores alongside their predictions, we can improve the precision of automatic error detection.
Leveraging confidence can reduce re-annotation efforts by flagging only cases exceeding a predefined threshold. The rationale is that not all flagged examples should be treated equally. Instances flagged with low confidence indicate that the LLM recognizes a potential issue, however, when the LLM is highly confident in a label that contradicts the original one, it provides a stronger signal of a possible error.

\begin{figure}[!t]
    \centering
    \includegraphics[width=0.99\linewidth]{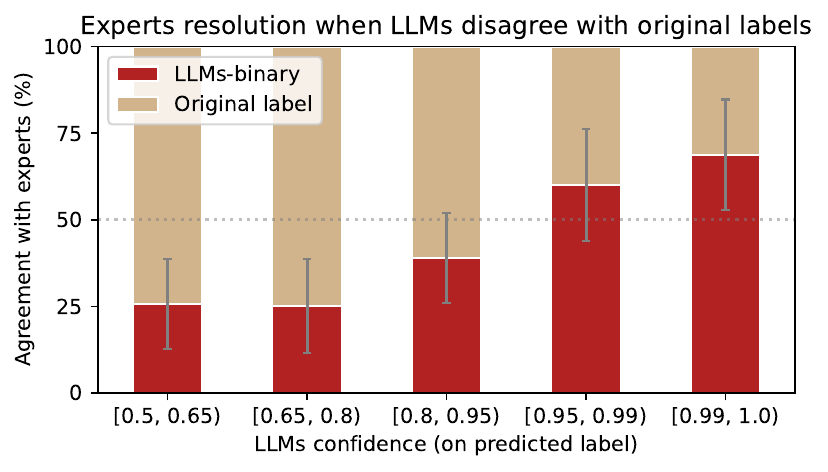}
    \includegraphics[width=0.99\linewidth]{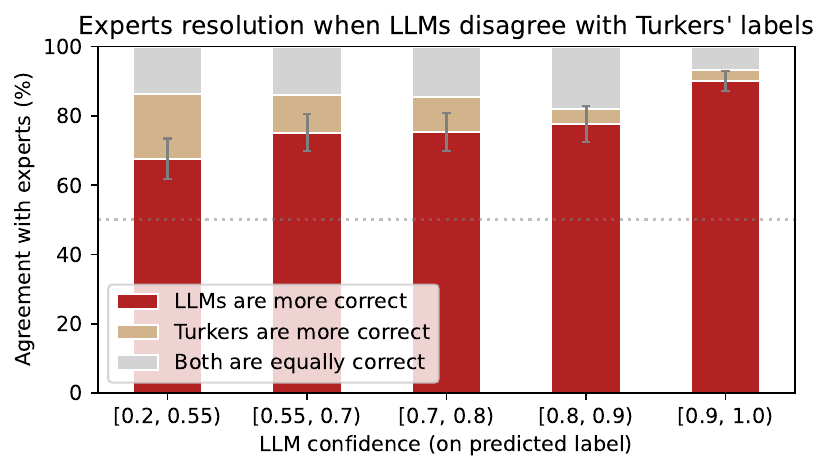}
    \caption{When LLMs disagree with original labels - who is correct? \textbf{(Top)} TRUE \textbf{(Bottom)} SummEval. As the LLM's confidence grows, so does the precision of identifying an error in the original labels.}
    \label{fig:experts_resolution}
    \vspace{-1em}
\end{figure}

\autoref{fig:experts_resolution} shows the rate of the experts' agreement with the LLMs compared to the agreement with original labels, divided into confidence-based bins. Bins are balanced by size, and defined by a confidence interval of 95\% based on bootstrap sampling (see Appendix \ref{sec: appendix-bootstrap} for further details). The bins reflect increasing levels of LLM confidence in its predicted label (i.e., a stronger disagreement between LLMs and the original labels).

From the top of \autoref{fig:experts_resolution}, we observe a clear trend: as LLM confidence increases, so does its precision in detecting label errors in the original dataset. In the highest confidence bin, LLM annotations surpass the original labels in agreement with expert re-labeling, and this difference is statistically significant. This indicates that when the LLM is highly confident in its disagreement with the original label, the labeled example serves as a strong candidate for a labeling error. Note that even in cases where the expert agreement with LLMs was below 50\%, mislabeled data was still discovered. See \autoref{sec:appendix_model_specific} for model-specific analysis.

We replicated this analysis on the SummEval dataset (bottom of \autoref{fig:experts_resolution}) and observed a similar trend: higher confidence increases the likelihood that the LLM prediction is closer to the expert annotation than the original label. In the SummEval case, we consider the crowd-sourced labels as the original labels. For more details see \autoref{sec: appendix_summeval}.

\begin{figure}[t]
    \centering
    \includegraphics[width=0.99\linewidth]{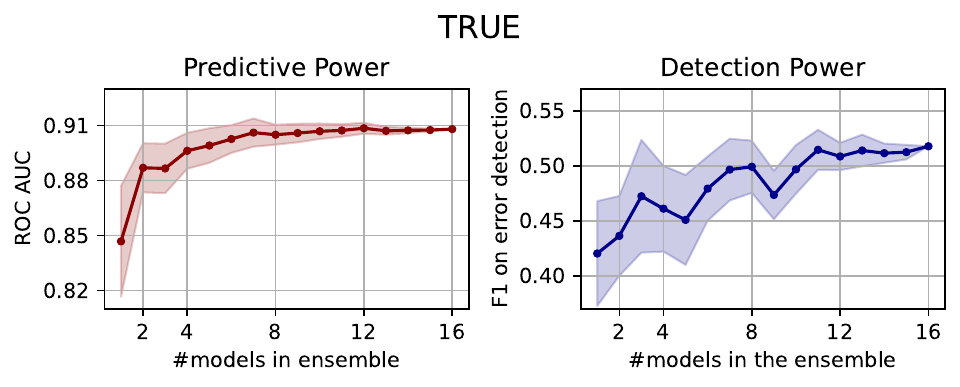}
    \includegraphics[width=0.99\linewidth]{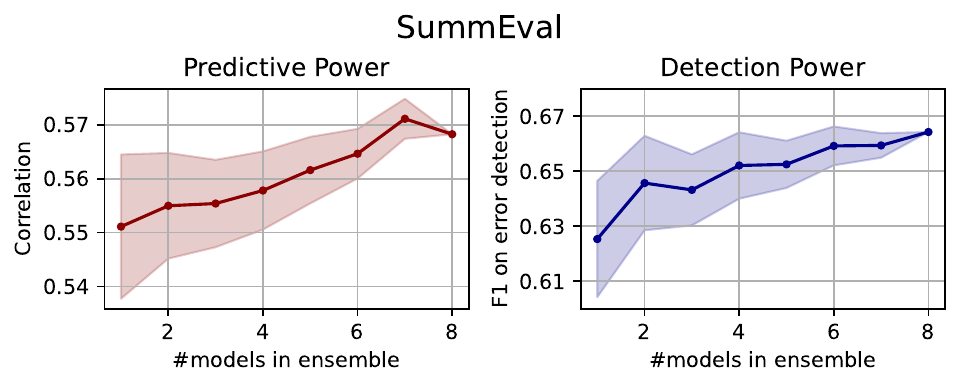}
    \caption{The power of ensemble. \textbf{(Top)} TRUE \textbf{(Bottom)} SummEval. As the ensemble size increases (\textbf{x-axis}), its performance against gold labels (\textbf{Left}), and its ability to detect label errors (\textbf{Right}) improve.}

    \label{fig:ensemble_main}
    \vspace{-1em}
\end{figure}

\paragraphn{Ensemble} By varying the size of the LLM ensemble, we examine two key aspects: predictive power (how well predictions align with gold labels, measured by ROC AUC for TRUE and average correlation for SummEval), and error detection power (measured by F1-score, averaging the recall of errors and the precision of correctly identifying a candidate as a true error). The ensemble power analysis is presented in \autoref{fig:ensemble_main}. 

For both aspects, we see a clear trend. As we increase the number of models in the ensemble, the performance increases. A higher ROC AUC with respect to the gold labels (left) reflects better annotation quality, while a higher F1 score (right) indicates a stronger error detector, either by recalling more errors or improving precision, or through a balance of both. Notably, to place the absolute F1-score in context, the expected F1-score for random behavior is approximately 0.22 (when randomly flagging errors), or around 0.13 (when randomly guessing the annotation), due to the class imbalance between error and non-error cases. Additionally, for both measures, the variance decreases as the ensemble size grows, which indicates more stable and consistent annotations and error detections. Similarly, \autoref{fig:ensemble_main} (bottom) shows the power of LLM ensemble on the same aspects on the SummEval datasets, aggregated over four summarization dimensions (see experiment details on Appendix \ref{sec:appendix_ensemble_summeval}). Trends of diminishing variance and increased performance and error detection are observed here as well.

Although not yet discussed in the context of error detection with LLMs, these results align with previous work showing the power of ensemble \citep{Dietterich2007EnsembleMI}. 

Our findings show that incorporating multiple LLMs and prompts in an ensemble is valuable: as the ensemble size increases, both label quality and error detection improve. These observations justify our choice to use an ensemble of models rather than a single one.

\begin{symbolfootnotes}
\begin{table*}[t]
\centering
\begin{adjustbox}{width=0.75\textwidth}
\begin{tabular}{l|c|c|c|p{6em}|c}
\toprule
\textbf{Annotator group} & \textbf{Fleiss's $\kappa$} & \textbf{\%agreement} &  \textbf{\#examples} & \textbf{Fleiss's $\kappa$ \newline (disagree. subset)} & \textbf{\#annotators} \\
\midrule
\textbf{Experts} & & & 222 & & 2 \\
\quad Before reconciliation & 0.486 & 75.7 & & \centering 0.486 & \\
\quad After reconciliation & 0.851 & 93.2 & & \centering 0.851 & \\
\midrule
\textbf{MTurk} & 0.074 & 60.5 & 400 & \centering -0.004 & 3\footnotemark[1] \\
\midrule
\textbf{LLM (different prompts)} & & & 640 &  & 4 \\
\quad GPT-4 & 0.706 & 85.3 & & \centering 0.571 & \\
\quad PaLM2 & 0.750 & 87.7 & & \centering 0.696 & \\
\quad LLaMA3 & 0.219 & 71.7 & & \centering 0.078 & \\
\quad Mistral & 0.459 & 73.2 & & \centering 0.314 & \\
\midrule
\textbf{LLMs (different models)} & 0.521 & 77.5 & 640 & \centering 0.389 & 4 \\
\bottomrule
\end{tabular}
\end{adjustbox}
\caption{Inter-annotator agreement (IAA) across annotator groups. LLMs such as GPT-4 and PaLM2 approach expert-level agreement, while MTurk workers show low and inconsistent reliability. Results for SummEval are provided in \autoref{table: IAA_summeval} in the appendix.}
\label{table: IAA}
\end{table*}
\footnotetext[1]{Multiple MTurk workers have participated in annotation, yet exactly 3 annotations per example were obtained. Annotator independence assumption was made to calculate Fleiss's $\kappa$ as with 3 annotators.}
\end{symbolfootnotes}

\section{Comparing Annotation Approaches}
\label{sec: comparison}

\begin{figure}[t]
\centering
    \includegraphics[width=0.85\linewidth]{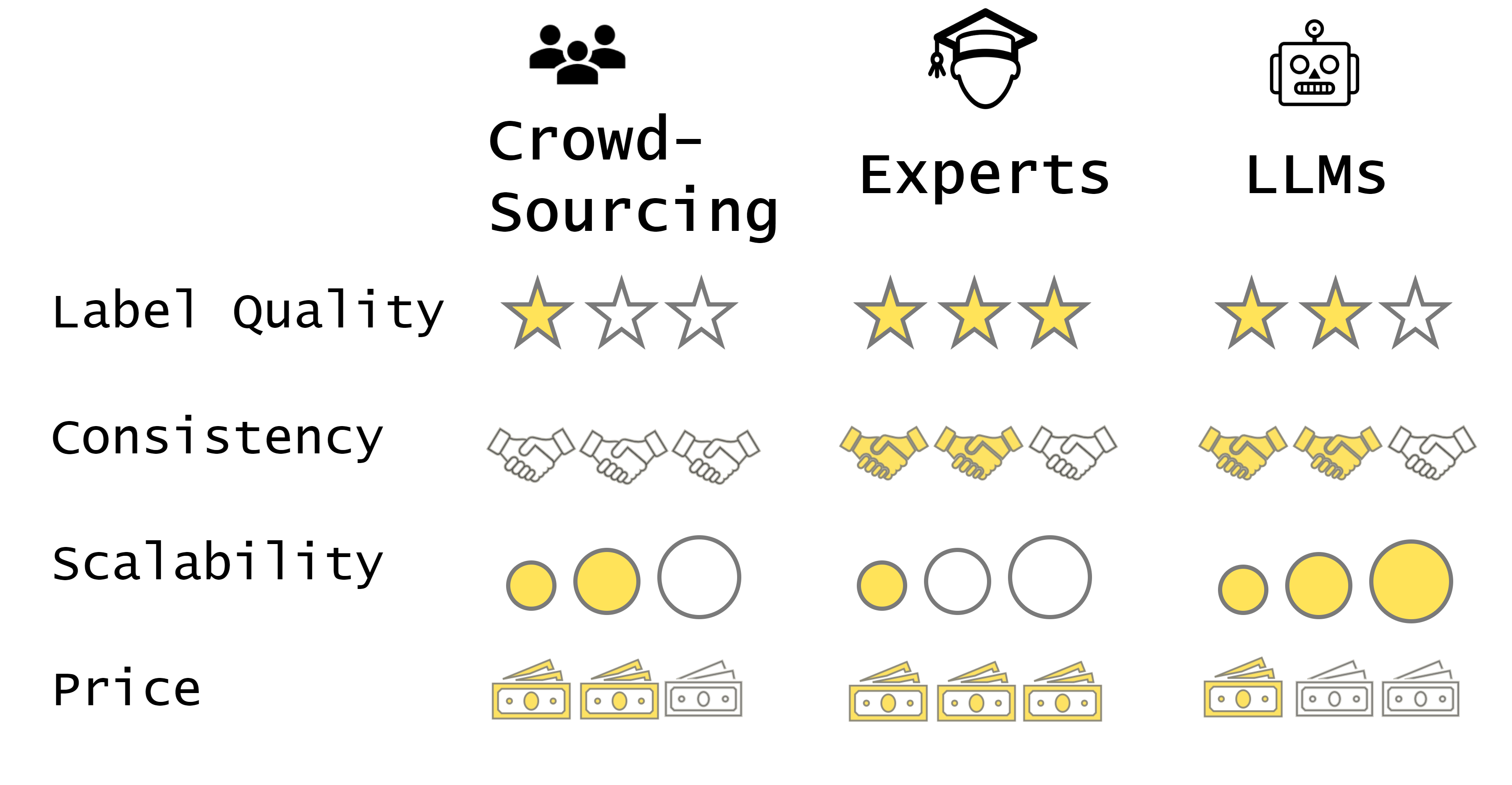}
    \caption{Annotation approaches comparison.}
    \label{fig:compare-dense}
    \vspace{-1em}
\end{figure}

Our paper discusses three annotation approaches, each with its own benefits and drawbacks, differing in how they balance label quality, scalability, and cost. Here we summarize the main findings, with additional analyses provided in \autoref{sec: appendix_comparison}. \autoref{fig:compare-dense} highlights the key results.  

\medskip\noindent\textbf{LLMs exhibit strong agreement with experts and among themselves.} Inter-annotator agreement (IAA) among LLMs, as well as their alignment with expert annotations, are significantly higher than that of crowd workers. As shown in \autoref{table: IAA}, GPT-4 and PaLM2 achieve $\kappa$ scores above 0.70, approaching expert-level agreement after reconciliation ($\kappa = 0.85$). In contrast, MTurk workers reach only $\kappa = 0.07$, underscoring the gap between crowd- and LLM-based annotation.  

\begin{figure}[t]
    \centering
    \includegraphics[width=0.95\linewidth]{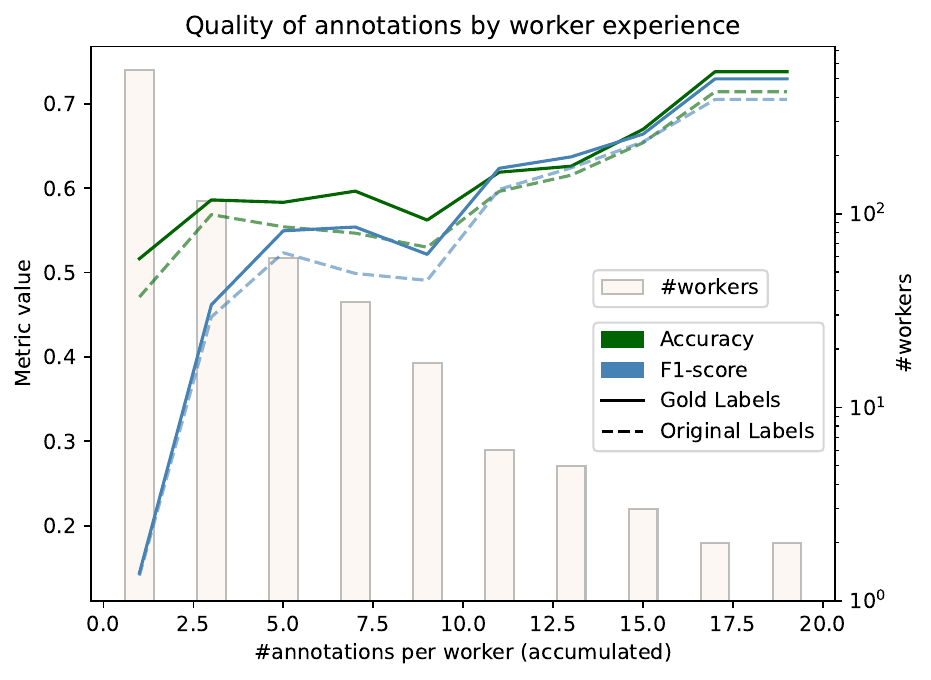}
    \caption{\textbf{(x-axis)} at list $x$ annotations per annotator. \textbf{(Right y-axis)} The number of annotators with at least $x$ annotations (bins). \textbf{(Left y-axis)} the average F1-score or accuracy for all user annotations with at least $x$ annotations.}
    \label{fig:mturk_experience-quality}
\end{figure}

\medskip\noindent\textbf{Crowd worker quality improves with experience but remains inconsistent.} Our analysis shows that experienced crowd workers produce higher-quality annotations, as illustrated in \autoref{fig:mturk_experience-quality}. However, even among them, annotation quality and consistency remain lower than LLM-based annotation, which is more reliable. This is reflected in the wide variance of MTurk agreement (60.5\% overall, $\kappa = -0.004$ on disagreement cases), suggesting that crowd annotation requires substantial verification to ensure reliability.  

\medskip\noindent\textbf{LLMs provide fast, scalable, and cost-efficient annotation.} Compared to expert and crowd-sourced annotation, LLMs require less time and are much more cost-effective per annotation. As discussed in \autoref{sec: cost-scale}, LLM annotation is estimated to be 100-1000 times cheaper than human annotation. This makes them a viable alternative for large-scale annotation while effectively balancing the trade-off.

\section{Implications of Mislabeled Data}
\label{sec: implications}

\subsection{Training on Mislabeled Data}
\label{sec: implications-training}

Training on mislabeled data can harm model performance and stability, as learning from errors makes it harder to identify consistent patterns. The impact depends on various factors, such as the fraction of mislabeled data and the training procedure. In this subsection, we show that addressing this issue, even heuristically, significantly improves the model's performance on a test set.

\paragraphn{Handling Label Errors}
In order to handle label errors in the training set, and reduce its effect on model performance, we propose two manipulations. For both manipulations, we flag examples where the model strongly disagrees with the original label(i.e., with confidence above a certain threshold). The first manipulation is \textit{filtering} flagged examples out, which maintains a ``cleaner'' yet smaller training set. The second manipulation is label \textit{flipping} for flagged examples, which maintains the same amount of data, but may also cause harm if flipping too many correct labels.

\begin{figure}[!t]
    \centering
    \includegraphics[width=0.99\linewidth]{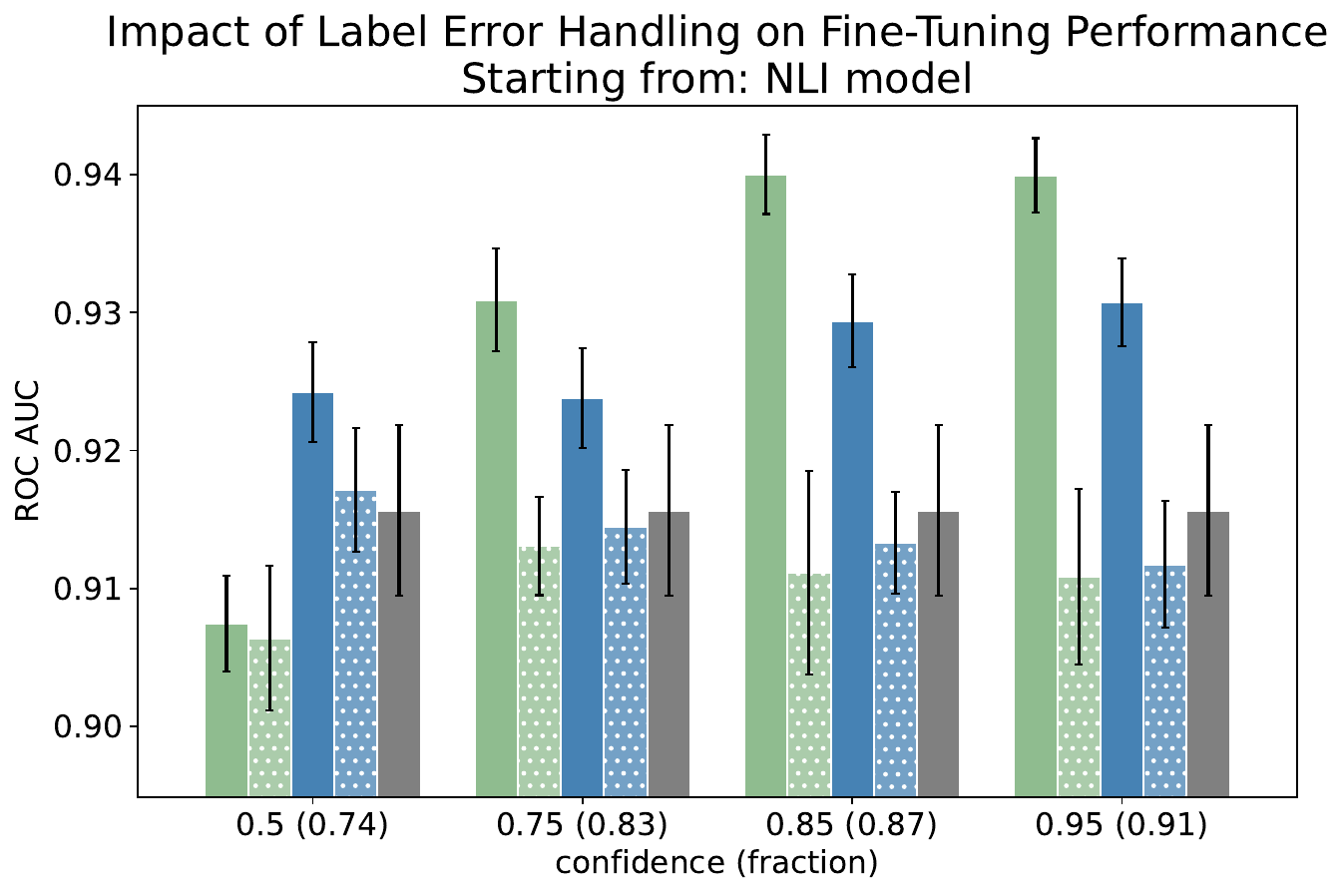}
    \includegraphics[width=0.9\linewidth]{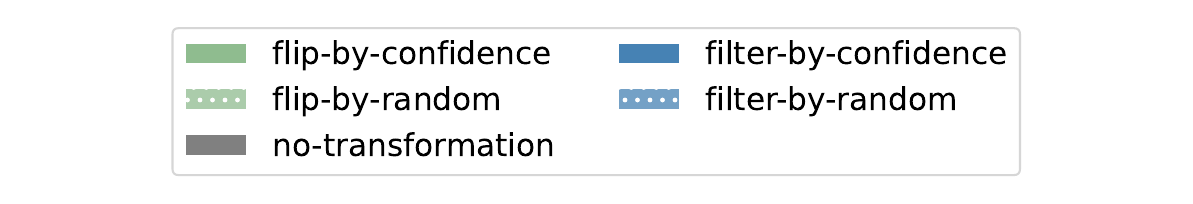}
    \caption{Fine-tuning a model on a transformed dataset. The gray bar is the original dataset - without any changes. The green bars present results for label flipping for a subset of examples, determined by LLMs-confidence (plain), or at random (dotted). The blue bars represent filtering of these examples.}
    \label{fig:finetuning}
    \vspace{-1em}
    
\end{figure}

\paragraphn{Experimental Setup}
We set the training set to be the additional data examples from the datasets (i.e., MNBM, BEGIN, VitaminC, PAWS), which are disjoint from the test set. Note that we posses gold labels for the test set alone, while for the training set we only extract the confidence. The finetuning procedure includes splitting the training set into train and validation sets, and fine-tuning on the train set. We report average results of five seeds.


As an ablation study, we also apply these manipulations on a random subset of examples rather than the flagged examples. 
The ablation study aims to maintain a consistent number of training examples, while the ablation for flipping aims to address the claim that in some cases, a relatively small fraction of label errors may be even considered as a noise that improves model robustness (e.g., as in label perturbation \citep{mixup} or label smoothing \citep{label_smoothing}).

We conducted this experiment starting from two base models: DeBERTa-v3, and a fine-tuned version of it on classic NLI datasets,  which we will refer to as the NLI-base model. We chose the NLI-base model as NLI tasks closely resemble factual consistency evaluation (FCE), making it well-suited for this experiment. Given the similar trends, we present the results for the NLI model here. Additional experiments and implementation details can be found in Appendix \ref{appendix-finetuning}.


\paragraphn{Results}
\autoref{fig:finetuning} shows the results of our experiments.
In our confidence-based approaches, we clearly see the trend that as the confidence threshold, according to which our manipulations are applied, grows, our manipulation results in improved ROC AUC for both models. This trend eventually (i.e., for high enough LLM confidence) brings these approaches to significantly outperform the baseline. In contrast, when we applied our manipulations on random subsets, we generally see a diminishing effect of manipulation, converging to the no-manipulation baseline.

Comparing between the handling approaches, it appears that flipping is better than filtering for high confidence. We hypothesize that this stems from the amount of data that remains after flipping (i.e., the same amount as before the flipping) compared to the filtering approach, combined with the high error rate in these datasets. Note that this is contrary to the random case where filtering is better than flipping, as flipping a subset with low error-rate brings more damage than value.

\subsection{Evaluating on Mislabeled Data}
\label{sec: implications-eval}

\begin{table*}[!t]
\centering

\begin{adjustbox}{width=0.85\textwidth}
\begin{tabular}{l|cl|cl|cl|cl}
\toprule
\multirow{2}{*}{\textbf{Model}} & \multicolumn{2}{c|}{\textbf{Rank}} & \multicolumn{2}{c|}{\textbf{ROC AUC}} & \multicolumn{2}{c|}{\textbf{F1 Score}} & \multicolumn{2}{c}{\textbf{Accuracy}} \\
  & Original & Gold & Original & Gold & Original & Gold & Original & Gold \\
\midrule
       GPT-4 &   3 &     1 (\textcolor{Green}{+2}) &  0.81 &   0.93 (\textcolor{Green}{+15\%}) & 0.73 &   0.83 (\textcolor{Green}{+14\%}) &  0.73 &   0.83 (\textcolor{Green}{+14\%}) \\
   NLI model &   1 & 2 (\textcolor{BrickRed}{--1}) &  0.93 & 0.91 (\textcolor{BrickRed}{--2\%}) & 0.87 &  0.87 (---) &  0.87 &  0.87 (---) \\
       PaLM2 &   6 &     3 (\textcolor{Green}{+3}) &  0.81 &   0.91 (\textcolor{Green}{+12\%}) & 0.71 &   0.81 (\textcolor{Green}{+14\%}) &  0.71 &   0.81 (\textcolor{Green}{+14\%}) \\
      GPT-4o &   4 &  4 (---) &  0.81 &   0.91 (\textcolor{Green}{+12\%}) & 0.74 &   0.83 (\textcolor{Green}{+12\%}) &  0.74 &   0.83 (\textcolor{Green}{+12\%}) \\
  GPT-4-mini &   5 &  5 (---) &  0.81 &   0.91 (\textcolor{Green}{+12\%}) & 0.71 &   0.79 (\textcolor{Green}{+11\%}) &  0.70 &   0.79 (\textcolor{Green}{+13\%}) \\
      Llama3 &   7 &     6 (\textcolor{Green}{+1}) &  0.75 &   0.86 (\textcolor{Green}{+15\%}) & 0.47 &     0.50 (\textcolor{Green}{+6\%}) &  0.52 &    0.55 (\textcolor{Green}{+6\%}) \\
Mistral-v0.3 &   8 &     7 (\textcolor{Green}{+1}) &  0.75 &   0.85 (\textcolor{Green}{+13\%}) & 0.61 &   0.68 (\textcolor{Green}{+11\%}) &  0.62 &   0.68 (\textcolor{Green}{+10\%}) \\
  DeBERTa-v3 &   2 & 8 (\textcolor{BrickRed}{--6}) &  0.84 &  0.80 (\textcolor{BrickRed}{--5\%}) & 0.76 & 0.73 (\textcolor{BrickRed}{--4\%}) &  0.76 & 0.73 (\textcolor{BrickRed}{--4\%}) \\
Mistral-v0.2 &   9 &  9 (---) &  0.73 &   0.82 (\textcolor{Green}{+12\%}) & 0.66 &    0.72 (\textcolor{Green}{+9\%}) &  0.66 &    0.72 (\textcolor{Green}{+9\%}) \\

\bottomrule
\end{tabular}
\end{adjustbox}
\caption{Comparison of Model Performance on Original and Gold Labels. Ranking is defined over ROC AUC.}
\label{table: model_performance}
\vspace{-1em}
\end{table*}

In this subsection, we examine the impact of mislabeled data in evaluation sets and its potential to distort results. Labeling errors can mislead the evaluation process, resulting in inaccurate performance metrics and, in some cases, flawed model comparisons that lead to incorrect conclusions.

\paragraphn{Experimental Setup} 
To test this assumption, we evaluate the performance of nine models, mostly state-of-the-art LLMs, on the test datasets. We compare their performance between the \emph{original} labels, and the \emph{gold} labels. For LLMs, we used zero-shot prediction as described in \autoref{sec:llm_method}, and averaged over prompts. For DeBERTa-based models, we used the fine-tuned models from \autoref{sec: implications-training}, and averaged over seeds. 

\paragraphn{Results} 
Prior to this work, an evaluation of these models would induce the values and ranking as in \autoref{table: model_performance} under the \textit{Original} sub-columns.
However, as shown before, these datasets include labeling errors, and therefore do not support fair evaluation. Considering the new gold labels, based on expert intervention (as described in \autoref{sec: annotation-expert}), we obtain different results, shown in the \textit{Gold} sub-columns.
The first observed discrepancy is the ranking of models. For example, DeBERTa-v3 has shifted from being the second-best to the second-worst.
Beyond the change in ranking, all metrics' (i.e., ROC AUC, F1-score, and accuracy) range has shifted upward, indicating that LLMs perform better on this task than previously thought. 
\draft{We further discuss the performance differences between LLMs and fine-tuned models in Appendix \ref{sec: appendix-eval}.}
If this phenomenon extends to other tasks and datasets beyond those examined in this study, it could suggest that LLMs are better than currently perceived.

\section{Discussion}



Labeling errors are a persistent issue in NLP datasets, negatively affecting model fine-tuning and evaluation. Our findings demonstrate that LLMs, particularly when highly confident, can effectively detect these errors, outperforming crowd workers in accuracy, consistency, and cost-efficiency. As LLM capabilities advance, their role in refining data quality will become central to improving NLP benchmarks. Future work could explore applying LLM-based error detection to a broader range of datasets and tasks, as well as refining methods for optimizing label correction strategies. We encourage researchers to adopt our methods and critically evaluate existing datasets to drive more robust, reliable results in the field.

\subsubsection*{Acknowledgements}
This research is a collaboration between the Technion and Google Research, supported by the Google Cloud Research Credits program with the award GCP19980904.  

\section*{Limitations}
While our study provides valuable insights into the role of LLMs in identifying label errors and improving dataset quality, several limitations should be considered.
First, crowd workers encompass a broad range of annotators with varying expertise and training.  Our analysis, focuses on the ``common'' crowd worker, typically an annotator selected with minimal qualifications, such as an approved task completion rate, and without specialized training. 
\draft{However, some datasets implement more selective strategies, such as requiring prior experience or task-specific training, which may yield more reliable labels. These "trained" crowd workers can be seen as an intermediate category between common annotators and experts, both in terms of cost and label quality. We chose to focus on the two endpoints, comparing common crowd workers and experts, to highlight clear contrasts in annotation quality and associated trade-offs.}
Importantly, we did not take crowd-worker annotations at face value; we applied filtering (based on the explanation crowd workers were asked to write for each example) to remove a substantial number of low-quality assignments, such as clearly invalid responses, in addition to enforcing minimal qualification criteria.

Second, our analysis does not account for potential data contamination, where LLMs may have been trained on the datasets we evaluate. However, since our analysis focuses on identifying and correcting label errors within these datasets, contamination would likely hinder rather than enhance our findings. If an LLM had memorized these datasets, it would be more likely to reproduce existing errors rather than detect and correct them, making contamination a potential limitation only for certain aspects of evaluation but not for our core claims.

Third, LLM-based annotations can vary depending on the choice of prompting strategies and ensemble methods. In this work, we use zero-shot prompting and simple averaging for ensembling. Still, alternative approaches -- such as few-shot prompting, chain-of-thought reasoning \citep{CoT}, or self-refine \citep{self_refine} -- could improve annotation accuracy and consistency. Likewise, for ensembling, more advanced methods- such as percentile-based aggregation \citep{10.7554/eLife.81916}, error-aware weighting \citep{adaboost}, confidence-aware methods \citep{confidence_ensemble, DBLP:conf/icml/LuB0XW24}, or even LLM-based aggregation strategies like debate variants \citep{DBLP:journals/corr/abs-2305-19118, multi-agent_debate} -- may yield more reliable consensus labels. We leave the exploration of these strategies for future work and hope our study encourages such further research.

\draft{Finally, while our study does not cover the full range of NLP tasks, it is grounded in diverse and realistic labeling settings. The TRUE benchmark includes factual consistency annotations for summarization, dialogue, paraphrasing, and fact verification. SummEval adds ordinal labels and evaluates multiple dimensions of summary quality, such as fluency and coherence. These datasets differ in task framing, label format, and domain, providing a solid basis for analyzing label errors and their effects. Extending this analysis to other task types is a valuable direction for future work.}

\section*{Ethical Considerations}
We address several ethical considerations related to human annotators and the research community.

First, we recognize the significant human effort and cost involved in creating the datasets used in this study. While we question certain labels in these datasets, this should not be seen as undermining their value or the hard work behind them. These datasets have been highly beneficial to the research community, and our aim is to help improve labeling quality, especially as powerful tools like LLMs become more capable in various tasks. Our goal is to highlight areas where improvements can be made, contributing to further advancements in the field.

Additionally, we used crowd-sourced human annotators for text labeling. All participants were paid fairly, in line with platform regulations and our institution’s policies. We ensured transparency in the process, treated participants with respect, and provided appropriate compensation for their efforts.

Lastly, we acknowledge the potential impact of LLMs on crowd-sourced workers who depend on these platforms for income. While we explore the use of LLMs to enhance or potentially replace certain aspects of annotation, we do not intend for this to harm human workers. Instead, we hope that crowd-sourced workers will adopt these tools, allowing them to become more efficient and skilled, which will improve both the scalability and quality of future datasets while maintaining a role for human oversight.



\bibliography{bibliography}

\begin{thebibliography}{89}
\providecommand{\natexlab}[1]{#1}

\bibitem[{Allahbakhsh et~al.(2013)Allahbakhsh, Benatallah, Ignjatovic, Motahari-Nezhad, Bertino, and Dustdar}]{Quality-Control-in-Crowdsourcing-issues}
Mohammad Allahbakhsh, Boualem Benatallah, Aleksandar Ignjatovic, Hamid~Reza Motahari-Nezhad, Elisa Bertino, and Schahram Dustdar. 2013.
\newblock \href {https://doi.org/10.1109/MIC.2013.20} {Quality control in crowdsourcing systems: Issues and directions}.
\newblock \emph{IEEE Internet Computing}, 17(2):76--81.

\bibitem[{Anil et~al.(2023)Anil, Dai, Firat, Johnson, Lepikhin, Passos, Shakeri, Taropa, Bailey, Chen, Chu, Clark, Shafey, Huang, Meier{-}Hellstern, Mishra, Moreira, Omernick, Robinson, Ruder, Tay, Xiao, Xu, Zhang, {\'{A}}brego, Ahn, Austin, Barham, Botha, Bradbury, Brahma, Brooks, Catasta, Cheng, Cherry, Choquette{-}Choo, Chowdhery, Crepy, Dave, Dehghani, Dev, Devlin, D{\'{\i}}az, Du, Dyer, Feinberg, Feng, Fienber, Freitag, Garcia, Gehrmann, Gonzalez, and et~al.}]{palm2}
Rohan Anil, Andrew~M. Dai, Orhan Firat, Melvin Johnson, Dmitry Lepikhin, Alexandre Passos, Siamak Shakeri, Emanuel Taropa, Paige Bailey, Zhifeng Chen, Eric Chu, Jonathan~H. Clark, Laurent~El Shafey, Yanping Huang, Kathy Meier{-}Hellstern, Gaurav Mishra, Erica Moreira, Mark Omernick, Kevin Robinson, Sebastian Ruder, Yi~Tay, Kefan Xiao, Yuanzhong Xu, Yujing Zhang, Gustavo~Hern{\'{a}}ndez {\'{A}}brego, Junwhan Ahn, Jacob Austin, Paul Barham, Jan~A. Botha, James Bradbury, Siddhartha Brahma, Kevin Brooks, Michele Catasta, Yong Cheng, Colin Cherry, Christopher~A. Choquette{-}Choo, Aakanksha Chowdhery, Cl{\'{e}}ment Crepy, Shachi Dave, Mostafa Dehghani, Sunipa Dev, Jacob Devlin, Mark D{\'{\i}}az, Nan Du, Ethan Dyer, Vladimir Feinberg, Fangxiaoyu Feng, Vlad Fienber, Markus Freitag, Xavier Garcia, Sebastian Gehrmann, Lucas Gonzalez, and et~al. 2023.
\newblock \href {https://doi.org/10.48550/ARXIV.2305.10403} {Palm 2 technical report}.
\newblock \emph{CoRR}, abs/2305.10403.

\bibitem[{Baledent et~al.(2022)Baledent, Mathet, Widl{\"o}cher, Couronne, and Manguin}]{Baledent2022ValidityAC}
Ana{\"e}lle Baledent, Yann Mathet, Antoine Widl{\"o}cher, Christophe Couronne, and Jean-Luc Manguin. 2022.
\newblock \href {https://api.semanticscholar.org/CorpusID:251465628} {Validity, agreement, consensuality and annotated data quality}.
\newblock In \emph{International Conference on Language Resources and Evaluation}.

\bibitem[{Bavaresco et~al.(2025)Bavaresco, Bernardi, Bertolazzi, Elliott, Fern{\'a}ndez, Gatt, Ghaleb, Giulianelli, Hanna, Koller, Martins, Mondorf, Neplenbroek, Pezzelle, Plank, Schlangen, Suglia, Surikuchi, Takmaz, and Testoni}]{bavaresco-etal-2025-llms}
Anna Bavaresco, Raffaella Bernardi, Leonardo Bertolazzi, Desmond Elliott, Raquel Fern{\'a}ndez, Albert Gatt, Esam Ghaleb, Mario Giulianelli, Michael Hanna, Alexander Koller, Andre Martins, Philipp Mondorf, Vera Neplenbroek, Sandro Pezzelle, Barbara Plank, David Schlangen, Alessandro Suglia, Aditya~K Surikuchi, Ece Takmaz, and Alberto Testoni. 2025.
\newblock \href {https://doi.org/10.18653/v1/2025.acl-short.20} {{LLM}s instead of human judges? a large scale empirical study across 20 {NLP} evaluation tasks}.
\newblock In \emph{Proceedings of the 63rd Annual Meeting of the Association for Computational Linguistics (Volume 2: Short Papers)}, pages 238--255, Vienna, Austria. Association for Computational Linguistics.

\bibitem[{Bhat and Varma(2023)}]{bhat-varma-2023-large}
Savita Bhat and Vasudeva Varma. 2023.
\newblock \href {https://doi.org/10.18653/v1/2023.eval4nlp-1.8} {Large language models as annotators: A preliminary evaluation for annotating low-resource language content}.
\newblock In \emph{Proceedings of the 4th Workshop on Evaluation and Comparison of NLP Systems}, pages 100--107, Bali, Indonesia. Association for Computational Linguistics.

\bibitem[{Brown et~al.(2020)Brown, Mann, Ryder, Subbiah, Kaplan, Dhariwal, Neelakantan, Shyam, Sastry, Askell, Agarwal, Herbert{-}Voss, Krueger, Henighan, Child, Ramesh, Ziegler, Wu, Winter, Hesse, Chen, Sigler, Litwin, Gray, Chess, Clark, Berner, McCandlish, Radford, Sutskever, and Amodei}]{GPT3}
Tom~B. Brown, Benjamin Mann, Nick Ryder, Melanie Subbiah, Jared Kaplan, Prafulla Dhariwal, Arvind Neelakantan, Pranav Shyam, Girish Sastry, Amanda Askell, Sandhini Agarwal, Ariel Herbert{-}Voss, Gretchen Krueger, Tom Henighan, Rewon Child, Aditya Ramesh, Daniel~M. Ziegler, Jeffrey Wu, Clemens Winter, Christopher Hesse, Mark Chen, Eric Sigler, Mateusz Litwin, Scott Gray, Benjamin Chess, Jack Clark, Christopher Berner, Sam McCandlish, Alec Radford, Ilya Sutskever, and Dario Amodei. 2020.
\newblock \href {https://proceedings.neurips.cc/paper/2020/hash/1457c0d6bfcb4967418bfb8ac142f64a-Abstract.html} {Language models are few-shot learners}.
\newblock In \emph{Advances in Neural Information Processing Systems 33: Annual Conference on Neural Information Processing Systems 2020, NeurIPS 2020, December 6-12, 2020, virtual}.

\bibitem[{Calderon et~al.(2024)Calderon, Porat, Ben-David, Chapanin, Gekhman, Oved, Shalumov, and Reichart}]{calderon2024measuring}
Nitay Calderon, Naveh Porat, Eyal Ben-David, Alexander Chapanin, Zorik Gekhman, Nadav Oved, Vitaly Shalumov, and Roi Reichart. 2024.
\newblock \href {https://doi.org/10.48550/arXiv.2306.00168} {Measuring the robustness of nlp models to domain shifts}.
\newblock \emph{arXiv preprint arXiv:2306.00168}.

\bibitem[{Calderon and Reichart(2024)}]{calderon2024behalf}
Nitay Calderon and Roi Reichart. 2024.
\newblock \href {https://doi.org/10.48550/ARXIV.2407.19200} {On behalf of the stakeholders: Trends in {NLP} model interpretability in the era of llms}.
\newblock \emph{CoRR}, abs/2407.19200.

\bibitem[{Chau et~al.(2020)Chau, Balaneshin, Liu, and Linda}]{Chau2020UnderstandingTT}
Hung Chau, Saeid Balaneshin, Kai Liu, and Ondrej Linda. 2020.
\newblock \href {https://api.semanticscholar.org/CorpusID:227231506} {Understanding the tradeoff between cost and quality of expert annotations for keyphrase extraction}.
\newblock In \emph{Law}.

\bibitem[{Chen and Ding(2023)}]{chen-ding-2023-probing}
Honghua Chen and Nai Ding. 2023.
\newblock \href {https://doi.org/10.18653/v1/2023.findings-emnlp.858} {Probing the {``}creativity{''} of large language models: Can models produce divergent semantic association?}
\newblock In \emph{Findings of the Association for Computational Linguistics: EMNLP 2023}, pages 12881--12888, Singapore. Association for Computational Linguistics.

\bibitem[{Chen et~al.(2024)Chen, Qin, Jiang, and Choi}]{Chen2024IsAL}
Ruirui Chen, Chengwei Qin, Weifeng Jiang, and Dongkyu Choi. 2024.
\newblock \href {https://api.semanticscholar.org/CorpusID:268710109} {Is a large language model a good annotator for event extraction?}
\newblock In \emph{AAAI Conference on Artificial Intelligence}.

\bibitem[{Chiang and Lee(2023)}]{chiang-lee-2023-large}
Cheng-Han Chiang and Hung-yi Lee. 2023.
\newblock \href {https://doi.org/10.18653/v1/2023.acl-long.870} {Can large language models be an alternative to human evaluations?}
\newblock In \emph{Proceedings of the 61st Annual Meeting of the Association for Computational Linguistics (Volume 1: Long Papers)}, pages 15607--15631, Toronto, Canada. Association for Computational Linguistics.

\bibitem[{Chmielewski and Kucker(2019)}]{Chmielewski2019AnMC}
Michael Chmielewski and Sarah~C. Kucker. 2019.
\newblock \href {https://api.semanticscholar.org/CorpusID:210355348} {An {MTurk} crisis? shifts in data quality and the impact on study results}.
\newblock \emph{Social Psychological and Personality Science}, 11:464 -- 473.

\bibitem[{Chong et~al.(2022)Chong, Hong, and Manning}]{ChongHM22}
Derek Chong, Jenny Hong, and Christopher~D. Manning. 2022.
\newblock \href {https://doi.org/10.18653/V1/2022.EMNLP-MAIN.618} {Detecting label errors by using pre-trained language models}.
\newblock In \emph{Proceedings of the 2022 Conference on Empirical Methods in Natural Language Processing, {EMNLP} 2022, Abu Dhabi, United Arab Emirates, December 7-11, 2022}, pages 9074--9091. Association for Computational Linguistics.

\bibitem[{Clopper and Pearson(1934)}]{clopper-pearson}
C.~J. Clopper and E.~S. Pearson. 1934.
\newblock \href {http://www.jstor.org/stable/2331986} {The use of confidence or fiducial limits illustrated in the case of the binomial}.
\newblock \emph{Biometrika}, 26(4):404--413.

\bibitem[{Devlin et~al.(2019)Devlin, Chang, Lee, and Toutanova}]{BERT}
Jacob Devlin, Ming{-}Wei Chang, Kenton Lee, and Kristina Toutanova. 2019.
\newblock \href {https://doi.org/10.18653/V1/N19-1423} {{BERT:} pre-training of deep bidirectional transformers for language understanding}.
\newblock In \emph{Proceedings of the 2019 Conference of the North American Chapter of the Association for Computational Linguistics: Human Language Technologies, {NAACL-HLT} 2019, Minneapolis, MN, USA, June 2-7, 2019, Volume 1 (Long and Short Papers)}, pages 4171--4186. Association for Computational Linguistics.

\bibitem[{Dietterich(2007)}]{Dietterich2007EnsembleMI}
Thomas~G. Dietterich. 2007.
\newblock \href {https://api.semanticscholar.org/CorpusID:10765854} {Ensemble methods in machine learning}.

\bibitem[{Dinan et~al.(2019)Dinan, Roller, Shuster, Fan, Auli, and Weston}]{WoW}
Emily Dinan, Stephen Roller, Kurt Shuster, Angela Fan, Michael Auli, and Jason Weston. 2019.
\newblock \href {https://openreview.net/forum?id=r1l73iRqKm} {Wizard of wikipedia: Knowledge-powered conversational agents}.
\newblock In \emph{7th International Conference on Learning Representations, {ICLR} 2019, New Orleans, LA, USA, May 6-9, 2019}. OpenReview.net.

\bibitem[{Du et~al.(2024)Du, Li, Torralba, Tenenbaum, and Mordatch}]{multi-agent_debate}
Yilun Du, Shuang Li, Antonio Torralba, Joshua~B. Tenenbaum, and Igor Mordatch. 2024.
\newblock \href {https://openreview.net/forum?id=zj7YuTE4t8} {Improving factuality and reasoning in language models through multiagent debate}.
\newblock In \emph{Forty-first International Conference on Machine Learning, {ICML} 2024, Vienna, Austria, July 21-27, 2024}. OpenReview.net.

\bibitem[{Dubey et~al.(2024)Dubey, Jauhri, Pandey, Kadian, Al{-}Dahle, Letman, and et~al.}]{llama3}
Abhimanyu Dubey, Abhinav Jauhri, Abhinav Pandey, Abhishek Kadian, Ahmad Al{-}Dahle, Aiesha Letman, and et~al. 2024.
\newblock \href {https://doi.org/10.48550/ARXIV.2407.21783} {The llama 3 herd of models}.
\newblock \emph{CoRR}, abs/2407.21783.

\bibitem[{Dziri et~al.(2022)Dziri, Rashkin, Linzen, and Reitter}]{BEGIN}
Nouha Dziri, Hannah Rashkin, Tal Linzen, and David Reitter. 2022.
\newblock \href {https://doi.org/10.1162/tacl_a_00506} {Evaluating attribution in dialogue systems: The {BEGIN} benchmark}.
\newblock \emph{Transactions of the Association for Computational Linguistics}, 10:1066--1083.

\bibitem[{Fabbri et~al.(2021)Fabbri, Kryscinski, McCann, Xiong, Socher, and Radev}]{summeval}
Alexander~R. Fabbri, Wojciech Kryscinski, Bryan McCann, Caiming Xiong, Richard Socher, and Dragomir~R. Radev. 2021.
\newblock \href {https://doi.org/10.1162/TACL\_A\_00373} {Summeval: Re-evaluating summarization evaluation}.
\newblock \emph{Trans. Assoc. Comput. Linguistics}, 9:391--409.

\bibitem[{Felkner et~al.(2024)Felkner, Thompson, and May}]{Felkner2024GPTIN}
Virginia~K. Felkner, Jennifer~A. Thompson, and Jonathan May. 2024.
\newblock \href {https://api.semanticscholar.org/CorpusID:270045683} {Gpt is not an annotator: The necessity of human annotation in fairness benchmark construction}.
\newblock \emph{ArXiv}, abs/2405.15760.

\bibitem[{Fleiss(1971)}]{Fleiss1971MeasuringNS}
Joseph~L. Fleiss. 1971.
\newblock \href {https://api.semanticscholar.org/CorpusID:143544759} {Measuring nominal scale agreement among many raters.}
\newblock \emph{Psychological Bulletin}, 76:378--382.

\bibitem[{Fr{\'e}nay and Verleysen(2014)}]{Frnay2014ClassificationIT}
Beno{\^i}t Fr{\'e}nay and Michel Verleysen. 2014.
\newblock \href {https://api.semanticscholar.org/CorpusID:6054025} {Classification in the presence of label noise: A survey}.
\newblock \emph{IEEE Transactions on Neural Networks and Learning Systems}, 25:845--869.

\bibitem[{Freund and Schapire(1997)}]{adaboost}
Yoav Freund and Robert~E Schapire. 1997.
\newblock \href {https://doi.org/10.1006/jcss.1997.1504} {A decision-theoretic generalization of on-line learning and an application to boosting}.
\newblock \emph{Journal of Computer and System Sciences}, 55(1):119--139.

\bibitem[{Gat et~al.(2024)Gat, Calderon, Feder, Chapanin, Sharma, and Reichart}]{GatCFCSR24}
Yair~Ori Gat, Nitay Calderon, Amir Feder, Alexander Chapanin, Amit Sharma, and Roi Reichart. 2024.
\newblock \href {https://openreview.net/forum?id=UMfcdRIotC} {Faithful explanations of black-box {NLP} models using {LLM}-generated counterfactuals}.
\newblock In \emph{The Twelfth International Conference on Learning Representations, {ICLR} 2024, Vienna, Austria, May 7-11, 2024}. OpenReview.net.

\bibitem[{Gekhman et~al.(2023)Gekhman, Herzig, Aharoni, Elkind, and Szpektor}]{gekhman-etal-2023-trueteacher}
Zorik Gekhman, Jonathan Herzig, Roee Aharoni, Chen Elkind, and Idan Szpektor. 2023.
\newblock \href {https://doi.org/10.18653/v1/2023.emnlp-main.127} {{T}rue{T}eacher: Learning factual consistency evaluation with large language models}.
\newblock In \emph{Proceedings of the 2023 Conference on Empirical Methods in Natural Language Processing}, pages 2053--2070, Singapore. Association for Computational Linguistics.

\bibitem[{Gilardi et~al.(2023)Gilardi, Alizadeh, and Kubli}]{Gilardi2023ChatGPTOC}
Fabrizio Gilardi, Meysam Alizadeh, and Ma{\"e}l Kubli. 2023.
\newblock \href {https://api.semanticscholar.org/CorpusID:257766307} {{ChatGPT} outperforms crowd workers for text-annotation tasks}.
\newblock \emph{Proceedings of the National Academy of Sciences of the United States of America}, 120.

\bibitem[{Hao et~al.(2020)Hao, Zhang, Sumkin, Mohamed, and Wu}]{Hao2020InaccurateLI}
Degan Hao, Lei Zhang, Jules~H. Sumkin, Aly~A. Mohamed, and Shandong Wu. 2020.
\newblock \href {https://api.semanticscholar.org/CorpusID:211232156} {Inaccurate labels in weakly-supervised deep learning: Automatic identification and correction and their impact on classification performance}.
\newblock \emph{IEEE Journal of Biomedical and Health Informatics}, 24:2701--2710.

\bibitem[{Hauser et~al.(2021)Hauser, Moss, Rosenzweig, Jaffe, Robinson, and Litman}]{Hauser2021EvaluatingCA}
David~N. Hauser, Aaron~J. Moss, Cheskie Rosenzweig, Shalom~N. Jaffe, Jonathan Robinson, and Leib Litman. 2021.
\newblock \href {https://api.semanticscholar.org/CorpusID:240497347} {Evaluating {CloudResearch’s} approved group as a solution for problematic data quality on {MTurk}}.
\newblock \emph{Behavior Research Methods}, 55:3953 -- 3964.

\bibitem[{He et~al.(2023)He, Lin, Gong, Jin, Zhang, Lin, Jiao, Yiu, Duan, and Chen}]{He2023AnnoLLMML}
Xingwei He, Zheng-Wen Lin, Yeyun Gong, Alex Jin, Hang Zhang, Chen Lin, Jian Jiao, Siu~Ming Yiu, Nan Duan, and Weizhu Chen. 2023.
\newblock \href {https://api.semanticscholar.org/CorpusID:257805087} {{AnnoLLM}: Making large language models to be better crowdsourced annotators}.
\newblock In \emph{North American Chapter of the Association for Computational Linguistics}.

\bibitem[{Hendrycks et~al.(2021)Hendrycks, Burns, Basart, Zou, Mazeika, Song, and Steinhardt}]{MMLU}
Dan Hendrycks, Collin Burns, Steven Basart, Andy Zou, Mantas Mazeika, Dawn Song, and Jacob Steinhardt. 2021.
\newblock \href {https://openreview.net/forum?id=d7KBjmI3GmQ} {Measuring massive multitask language understanding}.
\newblock In \emph{9th International Conference on Learning Representations, {ICLR} 2021, Virtual Event, Austria, May 3-7, 2021}. OpenReview.net.

\bibitem[{Honovich et~al.(2022)Honovich, Aharoni, Herzig, Taitelbaum, Kukliansky, Cohen, Scialom, Szpektor, Hassidim, and Matias}]{TRUE}
Or~Honovich, Roee Aharoni, Jonathan Herzig, Hagai Taitelbaum, Doron Kukliansky, Vered Cohen, Thomas Scialom, Idan Szpektor, Avinatan Hassidim, and Yossi Matias. 2022.
\newblock \href {https://doi.org/10.18653/V1/2022.NAACL-MAIN.287} {{TRUE:} re-evaluating factual consistency evaluation}.
\newblock In \emph{Proceedings of the 2022 Conference of the North American Chapter of the Association for Computational Linguistics: Human Language Technologies, {NAACL} 2022, Seattle, WA, United States, July 10-15, 2022}, pages 3905--3920. Association for Computational Linguistics.

\bibitem[{Jiang et~al.(2023)Jiang, Sablayrolles, Mensch, Bamford, Chaplot, de~Las~Casas, Bressand, Lengyel, Lample, Saulnier, Lavaud, Lachaux, Stock, Scao, Lavril, Wang, Lacroix, and Sayed}]{mistral7b}
Albert~Q. Jiang, Alexandre Sablayrolles, Arthur Mensch, Chris Bamford, Devendra~Singh Chaplot, Diego de~Las~Casas, Florian Bressand, Gianna Lengyel, Guillaume Lample, Lucile Saulnier, L{\'{e}}lio~Renard Lavaud, Marie{-}Anne Lachaux, Pierre Stock, Teven~Le Scao, Thibaut Lavril, Thomas Wang, Timoth{\'{e}}e Lacroix, and William~El Sayed. 2023.
\newblock \href {https://doi.org/10.48550/ARXIV.2310.06825} {Mistral 7b}.
\newblock \emph{CoRR}, abs/2310.06825.

\bibitem[{Kaplan et~al.(2020)Kaplan, McCandlish, Henighan, Brown, Chess, Child, Gray, Radford, Wu, and Amodei}]{DBLP:journals/corr/abs-2001-08361}
Jared Kaplan, Sam McCandlish, Tom Henighan, Tom~B. Brown, Benjamin Chess, Rewon Child, Scott Gray, Alec Radford, Jeffrey Wu, and Dario Amodei. 2020.
\newblock \href {https://arxiv.org/abs/2001.08361} {Scaling laws for neural language models}.
\newblock \emph{CoRR}, abs/2001.08361.

\bibitem[{Kazai et~al.(2013)Kazai, Kamps, and Milic{-}Frayling}]{DBLP:journals/ir/KazaiKM13}
Gabriella Kazai, Jaap Kamps, and Natasa Milic{-}Frayling. 2013.
\newblock \href {https://doi.org/10.1007/S10791-012-9205-0} {An analysis of human factors and label accuracy in crowdsourcing relevance judgments}.
\newblock \emph{Inf. Retr.}, 16(2):138--178.

\bibitem[{Kennedy et~al.(2020)Kennedy, Clifford, Burleigh, Waggoner, Jewell, and Winter}]{kennedy2020shape}
Ryan Kennedy, Scott Clifford, Tyler Burleigh, Philip~D Waggoner, Ryan Jewell, and Nicholas~JG Winter. 2020.
\newblock \href {https://www.cambridge.org/core/journals/political-science-research-and-methods/article/shape-of-and-solutions-to-the-mturk-quality-crisis/521AEEB9A9753D5C6038440BD123826C} {The shape of and solutions to the {MTurk} quality crisis}.
\newblock \emph{Political Science Research and Methods}, 8(4):614--629.

\bibitem[{Kholodna et~al.(2024)Kholodna, Julka, Khodadadi, Gumus, and Granitzer}]{Kholodna2024LLMsIT}
Nataliia Kholodna, Sahib Julka, Mohammad Khodadadi, Muhammed~Nurullah Gumus, and Michael Granitzer. 2024.
\newblock \href {https://api.semanticscholar.org/CorpusID:268876095} {Llms in the loop: Leveraging large language model annotations for active learning in low-resource languages}.
\newblock \emph{ArXiv}, abs/2404.02261.

\bibitem[{Kim et~al.(2024)Kim, Mitra, Chen, Rahman, and Zhang}]{Kim2024MEGAnnoAH}
Han~Jun Kim, Kushan Mitra, Rafael~Li Chen, Sajjadur Rahman, and Dan Zhang. 2024.
\newblock \href {https://api.semanticscholar.org/CorpusID:268041346} {Meganno+: A human-llm collaborative annotation system}.
\newblock In \emph{Conference of the European Chapter of the Association for Computational Linguistics}.

\bibitem[{Klie et~al.(2023)Klie, Webber, and Gurevych}]{10.1162/coli_a_00464}
Jan-Christoph Klie, Bonnie Webber, and Iryna Gurevych. 2023.
\newblock \href {https://doi.org/10.1162/coli_a_00464} {Annotation error detection: Analyzing the past and present for a more coherent future}.
\newblock \emph{Computational Linguistics}, 49(1):157--198.

\bibitem[{Krippendorff(1970)}]{krippendorff1970reliability}
Klaus Krippendorff. 1970.
\newblock \href {https://doi.org/10.1177/001316447003000105} {Estimating the reliability, systematic error, and random error of interval data}.
\newblock \emph{Educational and Psychological Measurement}, 30(1):61--70.

\bibitem[{Lee(2010)}]{confidence_ensemble}
Chi{-}Hoon Lee. 2010.
\newblock \href {https://doi.org/10.1145/1835804.1835899} {Learning to combine discriminative classifiers: confidence based}.
\newblock In \emph{Proceedings of the 16th {ACM} {SIGKDD} International Conference on Knowledge Discovery and Data Mining, Washington, DC, USA, July 25-28, 2010}, pages 743--752. {ACM}.

\bibitem[{Li et~al.(2023)Li, Shi, Ziems, Kan, Chen, Liu, and Yang}]{Li2023CoAnnotatingUW}
Minzhi Li, Taiwei Shi, Caleb Ziems, Min-Yen Kan, Nancy~F. Chen, Zhengyuan Liu, and Diyi Yang. 2023.
\newblock \href {https://api.semanticscholar.org/CorpusID:264439555} {Coannotating: Uncertainty-guided work allocation between human and large language models for data annotation}.
\newblock \emph{ArXiv}, abs/2310.15638.

\bibitem[{Liang et~al.(2023)Liang, He, Jiao, Wang, Wang, Wang, Yang, Tu, and Shi}]{DBLP:journals/corr/abs-2305-19118}
Tian Liang, Zhiwei He, Wenxiang Jiao, Xing Wang, Yan Wang, Rui Wang, Yujiu Yang, Zhaopeng Tu, and Shuming Shi. 2023.
\newblock \href {https://doi.org/10.48550/ARXIV.2305.19118} {Encouraging divergent thinking in large language models through multi-agent debate}.
\newblock \emph{CoRR}, abs/2305.19118.

\bibitem[{Lissak et~al.(2024)Lissak, Calderon, Shenkman, Ophir, Fruchter, Klomek, and Reichart}]{LissakCSOFKR24}
Shir Lissak, Nitay Calderon, Geva Shenkman, Yaakov Ophir, Eyal Fruchter, Anat~Brunstein Klomek, and Roi Reichart. 2024.
\newblock \href {https://doi.org/10.18653/V1/2024.NAACL-LONG.113} {The colorful future of llms: Evaluating and improving llms as emotional supporters for queer youth}.
\newblock In \emph{Proceedings of the 2024 Conference of the North American Chapter of the Association for Computational Linguistics: Human Language Technologies (Volume 1: Long Papers), {NAACL} 2024, Mexico City, Mexico, June 16-21, 2024}, pages 2040--2079. Association for Computational Linguistics.

\bibitem[{Lu et~al.(2020)Lu, Li, Wang, and Zhang}]{Quality-Control-of-Crowdsourcing}
Jian Lu, Wei Li, Qingren Wang, and Yiwen Zhang. 2020.
\newblock \href {https://doi.org/10.1109/DASC-PICom-CBDCom-CyberSciTech49142.2020.00044} {Research on data quality control of crowdsourcing annotation: A survey}.
\newblock In \emph{2020 IEEE Intl Conf on Dependable, Autonomic and Secure Computing, Intl Conf on Pervasive Intelligence and Computing, Intl Conf on Cloud and Big Data Computing, Intl Conf on Cyber Science and Technology Congress (DASC/PiCom/CBDCom/CyberSciTech)}, pages 201--208.

\bibitem[{Lu et~al.(2024)Lu, Bai, Li, Xiao, and Wang}]{DBLP:conf/icml/LuB0XW24}
Zhihe Lu, Jiawang Bai, Xin Li, Zeyu Xiao, and Xinchao Wang. 2024.
\newblock \href {https://openreview.net/forum?id=Lc1HlMo77m} {Beyond sole strength: Customized ensembles for generalized vision-language models}.
\newblock In \emph{Forty-first International Conference on Machine Learning, {ICML} 2024, Vienna, Austria, July 21-27, 2024}. OpenReview.net.

\bibitem[{MacCartney and Manning(2009)}]{maccartney-manning-2009-extended}
Bill MacCartney and Christopher~D. Manning. 2009.
\newblock \href {https://aclanthology.org/W09-3714} {An extended model of natural logic}.
\newblock In \emph{Proceedings of the Eight International Conference on Computational Semantics}, pages 140--156, Tilburg, The Netherlands. Association for Computational Linguistics.

\bibitem[{Madaan et~al.(2023)Madaan, Tandon, Gupta, Hallinan, Gao, Wiegreffe, Alon, Dziri, Prabhumoye, Yang, Gupta, Majumder, Hermann, Welleck, Yazdanbakhsh, and Clark}]{self_refine}
Aman Madaan, Niket Tandon, Prakhar Gupta, Skyler Hallinan, Luyu Gao, Sarah Wiegreffe, Uri Alon, Nouha Dziri, Shrimai Prabhumoye, Yiming Yang, Shashank Gupta, Bodhisattwa~Prasad Majumder, Katherine Hermann, Sean Welleck, Amir Yazdanbakhsh, and Peter Clark. 2023.
\newblock \href {http://papers.nips.cc/paper\_files/paper/2023/hash/91edff07232fb1b55a505a9e9f6c0ff3-Abstract-Conference.html} {Self-refine: Iterative refinement with self-feedback}.
\newblock In \emph{Advances in Neural Information Processing Systems 36: Annual Conference on Neural Information Processing Systems 2023, NeurIPS 2023, New Orleans, LA, USA, December 10 - 16, 2023}.

\bibitem[{Maynez et~al.(2020)Maynez, Narayan, Bohnet, and McDonald}]{MNBM}
Joshua Maynez, Shashi Narayan, Bernd Bohnet, and Ryan~T. McDonald. 2020.
\newblock \href {https://doi.org/10.18653/V1/2020.ACL-MAIN.173} {On faithfulness and factuality in abstractive summarization}.
\newblock In \emph{Proceedings of the 58th Annual Meeting of the Association for Computational Linguistics, {ACL} 2020, Online, July 5-10, 2020}, pages 1906--1919. Association for Computational Linguistics.

\bibitem[{Narayan et~al.(2018)Narayan, Cohen, and Lapata}]{xsum}
Shashi Narayan, Shay~B. Cohen, and Mirella Lapata. 2018.
\newblock \href {https://doi.org/10.18653/v1/D18-1206} {Don{'}t give me the details, just the summary! topic-aware convolutional neural networks for extreme summarization}.
\newblock In \emph{Proceedings of the 2018 Conference on Empirical Methods in Natural Language Processing}, pages 1797--1807, Brussels, Belgium. Association for Computational Linguistics.

\bibitem[{Nguyen et~al.(2019)Nguyen, Mummadi, Ngo, Nguyen, Beggel, and Brox}]{Nguyen2019SELFLT}
Duc~Tam Nguyen, Chaithanya~Kumar Mummadi, Thi-Phuong-Nhung Ngo, Thi Hoai~Phuong Nguyen, Laura Beggel, and Thomas Brox. 2019.
\newblock \href {https://api.semanticscholar.org/CorpusID:203737303} {Self: Learning to filter noisy labels with self-ensembling}.
\newblock \emph{ArXiv}, abs/1910.01842.

\bibitem[{Northcutt et~al.(2021)Northcutt, Athalye, and Mueller}]{Northcutt2021PervasiveLE}
Curtis~G. Northcutt, Anish Athalye, and Jonas~W. Mueller. 2021.
\newblock \href {https://api.semanticscholar.org/CorpusID:232404905} {Pervasive label errors in test sets destabilize machine learning benchmarks}.
\newblock \emph{ArXiv}, abs/2103.14749.

\bibitem[{Northcutt et~al.(2019)Northcutt, Jiang, and Chuang}]{Northcutt2019ConfidentLE}
Curtis~G. Northcutt, Lu~Jiang, and Isaac~L. Chuang. 2019.
\newblock \href {https://api.semanticscholar.org/CorpusID:207870256} {Confident learning: Estimating uncertainty in dataset labels}.
\newblock \emph{J. Artif. Intell. Res.}, 70:1373--1411.

\bibitem[{OpenAI(2023)}]{gpt4}
OpenAI. 2023.
\newblock \href {https://doi.org/10.48550/ARXIV.2303.08774} {{GPT-4} technical report}.
\newblock \emph{CoRR}, abs/2303.08774.

\bibitem[{Ouyang et~al.(2022)Ouyang, Wu, Jiang, Almeida, Wainwright, Mishkin, Zhang, Agarwal, Slama, Ray, Schulman, Hilton, Kelton, Miller, Simens, Askell, Welinder, Christiano, Leike, and Lowe}]{DBLP:conf/nips/Ouyang0JAWMZASR22}
Long Ouyang, Jeffrey Wu, Xu~Jiang, Diogo Almeida, Carroll~L. Wainwright, Pamela Mishkin, Chong Zhang, Sandhini Agarwal, Katarina Slama, Alex Ray, John Schulman, Jacob Hilton, Fraser Kelton, Luke Miller, Maddie Simens, Amanda Askell, Peter Welinder, Paul~F. Christiano, Jan Leike, and Ryan Lowe. 2022.
\newblock \href {http://papers.nips.cc/paper\_files/paper/2022/hash/b1efde53be364a73914f58805a001731-Abstract-Conference.html} {Training language models to follow instructions with human feedback}.
\newblock In \emph{Advances in Neural Information Processing Systems 35: Annual Conference on Neural Information Processing Systems 2022, NeurIPS 2022, New Orleans, LA, USA, November 28 - December 9, 2022}.

\bibitem[{Pleiss et~al.(2020)Pleiss, Zhang, Elenberg, and Weinberger}]{Pleiss2020IdentifyingMD}
Geoff Pleiss, Tianyi Zhang, Ethan~R. Elenberg, and Kilian~Q. Weinberger. 2020.
\newblock \href {https://api.semanticscholar.org/CorpusID:210932316} {Identifying mislabeled data using the area under the margin ranking}.
\newblock \emph{ArXiv}, abs/2001.10528.

\bibitem[{Rajpurkar et~al.(2016)Rajpurkar, Zhang, Lopyrev, and Liang}]{rajpurkar-etal-2016-squad}
Pranav Rajpurkar, Jian Zhang, Konstantin Lopyrev, and Percy Liang. 2016.
\newblock \href {https://doi.org/10.18653/v1/D16-1264} {{SQ}u{AD}: 100,000+ questions for machine comprehension of text}.
\newblock In \emph{Proceedings of the 2016 Conference on Empirical Methods in Natural Language Processing}, pages 2383--2392, Austin, Texas. Association for Computational Linguistics.

\bibitem[{Reiss et~al.(2020)Reiss, Xu, Cutler, Muthuraman, and Eichenberger}]{ReissXCME20}
Frederick Reiss, Hong Xu, Bryan Cutler, Karthik Muthuraman, and Zachary Eichenberger. 2020.
\newblock \href {https://doi.org/10.18653/V1/2020.CONLL-1.16} {Identifying incorrect labels in the conll-2003 corpus}.
\newblock In \emph{Proceedings of the 24th Conference on Computational Natural Language Learning, CoNLL 2020, Online, November 19-20, 2020}, pages 215--226. Association for Computational Linguistics.

\bibitem[{Rogers et~al.(2013)Rogers, Sleeman, and Kinsella}]{RogersSK13}
Simon Rogers, Derek~H. Sleeman, and John Kinsella. 2013.
\newblock \href {https://doi.org/10.1109/JBHI.2013.2252182} {Investigating the disagreement between clinicians' ratings of patients in icus}.
\newblock \emph{{IEEE} J. Biomed. Health Informatics}, 17(4):843--852.

\bibitem[{Schuster et~al.(2021)Schuster, Fisch, and Barzilay}]{vitc}
Tal Schuster, Adam Fisch, and Regina Barzilay. 2021.
\newblock \href {https://doi.org/10.18653/v1/2021.naacl-main.52} {Get your vitamin {C}! robust fact verification with contrastive evidence}.
\newblock In \emph{Proceedings of the 2021 Conference of the North American Chapter of the Association for Computational Linguistics: Human Language Technologies}, pages 624--643, Online. Association for Computational Linguistics.

\bibitem[{Sherratt et~al.(2023)Sherratt, Gruson, Grah, Johnson, Niehus, Prasse, and et~al.}]{10.7554/eLife.81916}
Katharine Sherratt, Hugo Gruson, Rok Grah, Helen Johnson, Rene Niehus, Bastian Prasse, and et~al. 2023.
\newblock \href {https://doi.org/10.7554/eLife.81916} {Predictive performance of multi-model ensemble forecasts of covid-19 across european nations}.
\newblock \emph{eLife}, 12:e81916.

\bibitem[{Snow et~al.(2008)Snow, O'Connor, Jurafsky, and Ng}]{Snow2008CheapAF}
Rion Snow, Brendan~T. O'Connor, Dan Jurafsky, and A.~Ng. 2008.
\newblock \href {https://api.semanticscholar.org/CorpusID:7008675} {Cheap and fast – but is it good? evaluating non-expert annotations for natural language tasks}.
\newblock In \emph{Conference on Empirical Methods in Natural Language Processing}.

\bibitem[{Srivastava et~al.(2023)Srivastava, Rastogi, Rao, Shoeb, Abid, Fisch, Brown, Santoro, Gupta, Garriga{-}Alonso, Kluska, Lewkowycz, Agarwal, Power, Ray, Warstadt, Kocurek, Safaya, Tazarv, Xiang, Parrish, Nie, Hussain, Askell, Dsouza, Slone, Rahane, Iyer, Andreassen, and et~al.}]{BIG-bench}
Aarohi Srivastava, Abhinav Rastogi, Abhishek Rao, Abu Awal~Md Shoeb, Abubakar Abid, Adam Fisch, Adam~R. Brown, Adam Santoro, Aditya Gupta, Adri{\`{a}} Garriga{-}Alonso, Agnieszka Kluska, Aitor Lewkowycz, Akshat Agarwal, Alethea Power, Alex Ray, Alex Warstadt, Alexander~W. Kocurek, Ali Safaya, Ali Tazarv, Alice Xiang, Alicia Parrish, Allen Nie, Aman Hussain, Amanda Askell, Amanda Dsouza, Ambrose Slone, Ameet Rahane, Anantharaman~S. Iyer, Anders Andreassen, and et~al. 2023.
\newblock \href {https://openreview.net/forum?id=uyTL5Bvosj} {Beyond the imitation game: Quantifying and extrapolating the capabilities of language models}.
\newblock \emph{Trans. Mach. Learn. Res.}, 2023.

\bibitem[{Steen et~al.(2023)Steen, Opitz, Frank, and Markert}]{steen-etal-2023-little}
Julius Steen, Juri Opitz, Anette Frank, and Katja Markert. 2023.
\newblock \href {https://doi.org/10.18653/v1/2023.acl-short.79} {With a little push, {NLI} models can robustly and efficiently predict faithfulness}.
\newblock In \emph{Proceedings of the 61st Annual Meeting of the Association for Computational Linguistics (Volume 2: Short Papers)}, pages 914--924, Toronto, Canada. Association for Computational Linguistics.

\bibitem[{Sylolypavan et~al.(2023)Sylolypavan, Sleeman, Wu, and Sim}]{SylolypavanSWS23}
Aneeta Sylolypavan, Derek~H. Sleeman, Honghan Wu, and Malcolm Sim. 2023.
\newblock \href {https://doi.org/10.1038/S41746-023-00773-3} {The impact of inconsistent human annotations on {AI} driven clinical decision making}.
\newblock \emph{npj Digit. Medicine}, 6.

\bibitem[{Szegedy et~al.(2016)Szegedy, Vanhoucke, Ioffe, Shlens, and Wojna}]{label_smoothing}
Christian Szegedy, Vincent Vanhoucke, Sergey Ioffe, Jonathon Shlens, and Zbigniew Wojna. 2016.
\newblock \href {https://doi.org/10.1109/CVPR.2016.308} {Rethinking the inception architecture for computer vision}.
\newblock In \emph{2016 {IEEE} Conference on Computer Vision and Pattern Recognition, {CVPR} 2016, Las Vegas, NV, USA, June 27-30, 2016}, pages 2818--2826. {IEEE} Computer Society.

\bibitem[{Tam et~al.(2023)Tam, Mascarenhas, Zhang, Kwan, Bansal, and Raffel}]{tam-etal-2023-evaluating}
Derek Tam, Anisha Mascarenhas, Shiyue Zhang, Sarah Kwan, Mohit Bansal, and Colin Raffel. 2023.
\newblock \href {https://doi.org/10.18653/v1/2023.findings-acl.322} {Evaluating the factual consistency of large language models through news summarization}.
\newblock In \emph{Findings of the Association for Computational Linguistics: ACL 2023}, pages 5220--5255, Toronto, Canada. Association for Computational Linguistics.

\bibitem[{Thorne et~al.(2018)Thorne, Vlachos, Christodoulopoulos, and Mittal}]{thorne-etal-2018-fever}
James Thorne, Andreas Vlachos, Christos Christodoulopoulos, and Arpit Mittal. 2018.
\newblock \href {https://doi.org/10.18653/v1/N18-1074} {{FEVER}: a large-scale dataset for fact extraction and {VER}ification}.
\newblock In \emph{Proceedings of the 2018 Conference of the North {A}merican Chapter of the Association for Computational Linguistics: Human Language Technologies, Volume 1 (Long Papers)}, pages 809--819, New Orleans, Louisiana. Association for Computational Linguistics.

\bibitem[{T{\"o}rnberg(2023)}]{Trnberg2023ChatGPT4OE}
Petter T{\"o}rnberg. 2023.
\newblock \href {https://api.semanticscholar.org/CorpusID:258108255} {{ChatGPT-4} outperforms experts and crowd workers in annotating political {Twitter} messages with zero-shot learning}.
\newblock \emph{ArXiv}, abs/2304.06588.

\bibitem[{Uma et~al.(2021)Uma, Fornaciari, Hovy, Paun, Plank, and Poesio}]{DBLP:journals/jair/UmaFHPPP21}
Alexandra Uma, Tommaso Fornaciari, Dirk Hovy, Silviu Paun, Barbara Plank, and Massimo Poesio. 2021.
\newblock \href {https://doi.org/10.1613/JAIR.1.12752} {Learning from disagreement: {A} survey}.
\newblock \emph{J. Artif. Intell. Res.}, 72:1385--1470.

\bibitem[{Ventura et~al.(2023)Ventura, Ben{-}David, Korhonen, and Reichart}]{Ventura2023navigating}
Mor Ventura, Eyal Ben{-}David, Anna Korhonen, and Roi Reichart. 2023.
\newblock \href {https://doi.org/10.48550/ARXIV.2310.01929} {Navigating cultural chasms: Exploring and unlocking the cultural {POV} of text-to-image models}.
\newblock \emph{CoRR}, abs/2310.01929.

\bibitem[{Veselovsky et~al.(2023{\natexlab{a}})Veselovsky, Ribeiro, Cozzolino, Gordon, Rothschild, and West}]{veselovsky2023prevalencepreventionlargelanguage}
Veniamin Veselovsky, Manoel~Horta Ribeiro, Philip Cozzolino, Andrew Gordon, David Rothschild, and Robert West. 2023{\natexlab{a}}.
\newblock \href {https://doi.org/10.48550/ARXIV.2310.15683} {Prevalence and prevention of large language model use in crowd work}.
\newblock \emph{CoRR}, abs/2310.15683.

\bibitem[{Veselovsky et~al.(2023{\natexlab{b}})Veselovsky, Ribeiro, and West}]{artificial_users}
Veniamin Veselovsky, Manoel~Horta Ribeiro, and Robert West. 2023{\natexlab{b}}.
\newblock \href {https://doi.org/10.48550/ARXIV.2306.07899} {Artificial artificial artificial intelligence: Crowd workers widely use large language models for text production tasks}.
\newblock \emph{CoRR}, abs/2306.07899.

\bibitem[{Wang et~al.(2019)Wang, Singh, Michael, Hill, Levy, and Bowman}]{GLUE}
Alex Wang, Amanpreet Singh, Julian Michael, Felix Hill, Omer Levy, and Samuel~R. Bowman. 2019.
\newblock \href {https://openreview.net/forum?id=rJ4km2R5t7} {{GLUE:} {A} multi-task benchmark and analysis platform for natural language understanding}.
\newblock In \emph{7th International Conference on Learning Representations, {ICLR} 2019, New Orleans, LA, USA, May 6-9, 2019}. OpenReview.net.

\bibitem[{Wang et~al.(2024)Wang, Kulkarni, and Qi}]{wang-etal-2024-less}
Tong Wang, Ninad Kulkarni, and Yanjun Qi. 2024.
\newblock \href {https://doi.org/10.18653/v1/2024.naacl-industry.27} {Less is more for improving automatic evaluation of factual consistency}.
\newblock In \emph{Proceedings of the 2024 Conference of the North American Chapter of the Association for Computational Linguistics: Human Language Technologies (Volume 6: Industry Track)}, pages 324--334, Mexico City, Mexico. Association for Computational Linguistics.

\bibitem[{Wang et~al.(2022)Wang, Mishra, Alipoormolabashi, Kordi, Mirzaei, Naik, Ashok, Dhanasekaran, Arunkumar, Stap, Pathak, Karamanolakis, Lai, Purohit, Mondal, Anderson, Kuznia, Doshi, Pal, Patel, Moradshahi, Parmar, Purohit, Varshney, Kaza, Verma, Puri, Karia, Doshi, Sampat, Mishra, Reddy~A, Patro, Dixit, and Shen}]{Super-NaturalInstructions}
Yizhong Wang, Swaroop Mishra, Pegah Alipoormolabashi, Yeganeh Kordi, Amirreza Mirzaei, Atharva Naik, Arjun Ashok, Arut~Selvan Dhanasekaran, Anjana Arunkumar, David Stap, Eshaan Pathak, Giannis Karamanolakis, Haizhi Lai, Ishan Purohit, Ishani Mondal, Jacob Anderson, Kirby Kuznia, Krima Doshi, Kuntal~Kumar Pal, Maitreya Patel, Mehrad Moradshahi, Mihir Parmar, Mirali Purohit, Neeraj Varshney, Phani~Rohitha Kaza, Pulkit Verma, Ravsehaj~Singh Puri, Rushang Karia, Savan Doshi, Shailaja~Keyur Sampat, Siddhartha Mishra, Sujan Reddy~A, Sumanta Patro, Tanay Dixit, and Xudong Shen. 2022.
\newblock \href {https://doi.org/10.18653/v1/2022.emnlp-main.340} {Super-{N}atural{I}nstructions: Generalization via declarative instructions on 1600+ {NLP} tasks}.
\newblock In \emph{Proceedings of the 2022 Conference on Empirical Methods in Natural Language Processing}, pages 5085--5109, Abu Dhabi, United Arab Emirates. Association for Computational Linguistics.

\bibitem[{Weber and Plank(2023)}]{weber-plank-2023-activeaed}
Leon Weber and Barbara Plank. 2023.
\newblock \href {https://doi.org/10.18653/v1/2023.findings-acl.562} {{A}ctive{AED}: A human in the loop improves annotation error detection}.
\newblock In \emph{Findings of the Association for Computational Linguistics: ACL 2023}, pages 8834--8845, Toronto, Canada. Association for Computational Linguistics.

\bibitem[{Weber-Genzel et~al.(2024)Weber-Genzel, Peng, De~Marneffe, and Plank}]{weber-genzel-etal-2024-varierr}
Leon Weber-Genzel, Siyao Peng, Marie-Catherine De~Marneffe, and Barbara Plank. 2024.
\newblock \href {https://doi.org/10.18653/v1/2024.acl-long.123} {{V}ari{E}rr {NLI}: Separating annotation error from human label variation}.
\newblock In \emph{Proceedings of the 62nd Annual Meeting of the Association for Computational Linguistics (Volume 1: Long Papers)}, pages 2256--2269, Bangkok, Thailand. Association for Computational Linguistics.

\bibitem[{Wei et~al.(2022)Wei, Wang, Schuurmans, Bosma, Ichter, Xia, Chi, Le, and Zhou}]{CoT}
Jason Wei, Xuezhi Wang, Dale Schuurmans, Maarten Bosma, Brian Ichter, Fei Xia, Ed~H. Chi, Quoc~V. Le, and Denny Zhou. 2022.
\newblock \href {http://papers.nips.cc/paper\_files/paper/2022/hash/9d5609613524ecf4f15af0f7b31abca4-Abstract-Conference.html} {Chain-of-thought prompting elicits reasoning in large language models}.
\newblock In \emph{Advances in Neural Information Processing Systems 35: Annual Conference on Neural Information Processing Systems 2022, NeurIPS 2022, New Orleans, LA, USA, November 28 - December 9, 2022}.

\bibitem[{Williams et~al.(2018)Williams, Nangia, and Bowman}]{MNLI}
Adina Williams, Nikita Nangia, and Samuel Bowman. 2018.
\newblock \href {http://aclweb.org/anthology/N18-1101} {A broad-coverage challenge corpus for sentence understanding through inference}.
\newblock In \emph{Proceedings of the 2018 Conference of the North American Chapter of the Association for Computational Linguistics: Human Language Technologies, Volume 1 (Long Papers)}, pages 1112--1122. Association for Computational Linguistics.

\bibitem[{Xia et~al.(2012)Xia, Broadhurst, Wilson, and Wishart}]{Xia2012TranslationalBD}
Jianguo Xia, David~I. Broadhurst, Michael Wilson, and David~Scott Wishart. 2012.
\newblock \href {https://api.semanticscholar.org/CorpusID:7652615} {Translational biomarker discovery in clinical metabolomics: an introductory tutorial}.
\newblock \emph{Metabolomics}, 9:280 -- 299.

\bibitem[{Yang et~al.(2023)Yang, Mohamed, Jain, Peshterliev, Chatterjee, Zha, Bhalla, Aneja, and Mohanty}]{yang2023improvingopinionbasedquestionanswering}
Xiao Yang, Ahmed~K. Mohamed, Shashank Jain, Stanislav Peshterliev, Debojeet Chatterjee, Hanwen Zha, Nikita Bhalla, Gagan Aneja, and Pranab Mohanty. 2023.
\newblock \href {https://arxiv.org/abs/2306.07499} {Improving opinion-based question answering systems through label error detection and overwrite}.
\newblock \emph{Preprint}, arXiv:2306.07499.

\bibitem[{Zha et~al.(2023)Zha, Yang, Li, and Hu}]{Zha2023AlignScoreEF}
Yuheng Zha, Yichi Yang, Ruichen Li, and Zhiting Hu. 2023.
\newblock \href {https://api.semanticscholar.org/CorpusID:258947273} {Alignscore: Evaluating factual consistency with a unified alignment function}.
\newblock In \emph{Annual Meeting of the Association for Computational Linguistics}.

\bibitem[{Zhang et~al.(2018)Zhang, Ciss{\'{e}}, Dauphin, and Lopez{-}Paz}]{mixup}
Hongyi Zhang, Moustapha Ciss{\'{e}}, Yann~N. Dauphin, and David Lopez{-}Paz. 2018.
\newblock \href {https://openreview.net/forum?id=r1Ddp1-Rb} {mixup: Beyond empirical risk minimization}.
\newblock In \emph{6th International Conference on Learning Representations, {ICLR} 2018, Vancouver, BC, Canada, April 30 - May 3, 2018, Conference Track Proceedings}. OpenReview.net.

\bibitem[{Zhang et~al.(2023)Zhang, Li, Ma, Zhou, and Zou}]{Zhang2023LLMaAAML}
Ruoyu Zhang, Yanzeng Li, Yongliang Ma, Ming Zhou, and Lei Zou. 2023.
\newblock \href {https://api.semanticscholar.org/CorpusID:264814421} {{LLMaAA}: Making large language models as active annotators}.
\newblock \emph{ArXiv}, abs/2310.19596.

\bibitem[{Zhang et~al.(2019)Zhang, Baldridge, and He}]{PAWS}
Yuan Zhang, Jason Baldridge, and Luheng He. 2019.
\newblock \href {https://doi.org/10.18653/v1/N19-1131} {{PAWS}: Paraphrase adversaries from word scrambling}.
\newblock In \emph{Proceedings of the 2019 Conference of the North {A}merican Chapter of the Association for Computational Linguistics: Human Language Technologies, Volume 1 (Long and Short Papers)}, pages 1298--1308, Minneapolis, Minnesota. Association for Computational Linguistics.

\bibitem[{Zheng et~al.(2023)Zheng, Chiang, Sheng, Zhuang, Wu, Zhuang, Lin, Li, Li, Xing, Zhang, Gonzalez, and Stoica}]{llm_as_a_judge}
Lianmin Zheng, Wei{-}Lin Chiang, Ying Sheng, Siyuan Zhuang, Zhanghao Wu, Yonghao Zhuang, Zi~Lin, Zhuohan Li, Dacheng Li, Eric~P. Xing, Hao Zhang, Joseph~E. Gonzalez, and Ion Stoica. 2023.
\newblock \href {http://papers.nips.cc/paper\_files/paper/2023/hash/91f18a1287b398d378ef22505bf41832-Abstract-Datasets\_and\_Benchmarks.html} {Judging llm-as-a-judge with mt-bench and chatbot arena}.
\newblock In \emph{Advances in Neural Information Processing Systems 36: Annual Conference on Neural Information Processing Systems 2023, NeurIPS 2023, New Orleans, LA, USA, December 10 - 16, 2023}.

\end{thebibliography}

\clearpage

\appendix

\renewcommand \thepart{}
\renewcommand \partname{}
\mtcsettitle{parttoc}{}
\addcontentsline{toc}{section}{Appendix} 
\part{Appendix} 
\parttoc 

\section{Additional Experiments - SummEval}
\label{sec: appendix_summeval}
In addition to the datasets from the TRUE benchmark, we replicate our experiments on another dataset with a different objective and a different labeling scheme, to strengthen our results and conclusions.

\subsection{Data}
SummEval \citep{summeval} is an extensive and commonly used summarization benchmark, evaluating the quality of multiple model-generated summarization outputs compared to a source CNN/DailyMail sources on four dimensions: coherence, relevance, consistency, and fluency. Each summarization is labeled on each dimension with five crowd-workers and three experts, enabling us to replicate some of the experiments without additional crowd-worker or expert annotation costs. The labeling schema is ordinal on a scale of 1 to 5 (higher is better). 
Note that this dataset does not have a singular gold-standard label per summarization, but rather a collection of annotations from experts and crowd-workers. Therefore, we will not claim to find label errors in this benchmark, but rather showcase our methodology as if the crowd-sourced annotations are the original labels for the dataset, and we have access to experts' annotations for gold-standard reference, to determine if the LLM was correct when flagging examples.

\subsection{Definitions}
To apply our methods for error detection via LLMs ensemble, we first define the following:

\paragraphs{Labels} We aggregate crowd-sourced annotations by their median, to construct a single original label on a scale of 1 to 5. Similarly, we take the median of the experts' annotations to be a single gold-standard label.

\paragraphs{A disagreement} We say that the LLM annotation \textit{disagrees} with the original label if there is a difference of more than 1 between the scores. \draft{Smaller differences (e.g., 4 vs. 5) may reflect natural variation in subjective interpretation rather than a labeling mistake, and are therefore not considered strong disagreements. In practice, using a threshold of 1 results in over 50\% of the dataset being flagged, making it difficult to isolate meaningful errors. We adopt this more conservative threshold to better reflect genuine annotation issues and reduce noise in our error detection process.}


\subsection{Experimental Setting}
\label{sec: appendix_summeval_experimental}
Similar to the description in \autoref{sec: annotation-llm}, we utilize two LLMs-- GPT-4o (\codefont{gpt-4o-2024-11-20}) and Gemini 1.5 Flash (\codefont{gemini-1.5-flash-002}). We constructed four prompts, differing by phrasing and compatible with the four prompt template structures used for the TRUE benchmark experiments. The answer to each query was a JSON format with 'Relevance', 'Coherence', 'Consistency', and 'Fluency' as its keys. The scores are integers on a scale of 1 to 5, as are the ratings in the SummEval dataset. We extract the probability of each score possible through the log-probs for each score token. 
Finally, we average all models' probabilities, to obtain an ensemble of LLMs, with $p$ being the distribution over the five possible scores.

\subsection{Experiments and Results}
\subsubsection{Can LLMs Detect Label Errors?}
We replicate the experiment described in \autoref{sec: RQ2} with the appropriate adjustment for the SummEval dataset, based on the definitions above. The result is shown in \autoref{fig:experts_resolution} (bottom). The plot presents the subset of examples where there was a disagreement between the crowd-sourced annotation and the LLMs' annotation. Each bin represents the confidence of the LLMs in their predicted label. As there are five ordinal categories, even if there was a disagreement between two annotations, they both might be "wrong", where the expert's answer is a third option. Therefore, to show clearer results, we do not resolve by experts "who is correct", but rather "who is more correct?". For completeness, we also provide the "both equally correct" option, for the case the expert's label is exactly in the middle, and none is "more correct" than the other.
The bins are relatively balanced in terms of the amount of examples per bin. Note that in contrast to the TRUE binary labeling scheme, where confidence 0.5 is the minimal threshold for an answer, here we start from 0.2.

From the results, we see a clear dominance of the LLM over the crowd-sourced annotations, for all confidence bins. This suggests that the LLMs not only \textit{detect} error by flagging possibly mislabeled data points, but also provide better answers, which can account for error \textit{correction}. Similar to the result on the TRUE benchmark, we observe a trend where as the LLMs' confidence increases, they are more correct, indicating that they find label errors with higher precision. However, in this dataset, the difference from the original labels (in this case, the MTurk labels) is even more apparent, and the LLMs are correct even when with lower confidence.

\subsubsection{The Power of Ensemble}
\label{sec:appendix_ensemble_summeval}
We analyze the importance of utilizing more than a single model and a single prompt on two dimensions - performance compared to the gold labels (the quality of the annotations we utilize), and error detection (the ability to identify errors more accurately). For performance evaluation on the ordinal labels, we report Pearson correlation; for error detection evaluation, we report the F1-score based on binary error/not-error classification. See results in \autoref{fig:ensemble_main} and discussion in \autoref{sec: appendix-ensemble}.

\subsubsection{Annotation Approaches Comparison}
In \autoref{sec: appendix_comparison}, we thoroughly discuss the comparison between the different annotation approaches. For SummEval, experts and crowd-sourced annotations are provided. Together with our LLM-ensemble annotations (as described in \autoref{sec: appendix_summeval_experimental}), we analyze and compare the annotation approaches in terms of quality (see \autoref{fig:all_comparison} (bottom)) and consistency (see \autoref{table: IAA_summeval}). To account for ordinal labels, we measure IAA via Krippendorff's $\alpha$ \citep{krippendorff1970reliability}.

\section{Comparing Annotation Approaches}
\label{sec: appendix_comparison}

Our paper discusses three annotation approaches, each with its own benefits and drawbacks. These approaches differ in how they manage the trade-offs between label quality, scalability, and cost. In the following section, we discuss and compare their characteristics. A summary of this comparison is given in \autoref{fig:compare-dense}.

\subsection{Annotation Quality}
\label{sec: compare_accuracy}
When annotating or validating a dataset, one of our main concerns is the quality of the labels, or in other words, establishing a reliable gold standard. However, each annotation approach produces different labels. To estimate the quality of these approaches, we measure the agreement between different annotations using the weighted F1-score (which accounts for both classes). Note that this metric is not symmetric, meaning that treating one annotation as the \emph{true} label and the other as the \emph{prediction}, or vice versa, can result in different scores.

\begin{figure}[t]
    \centering
    \includegraphics[width=0.95\linewidth]{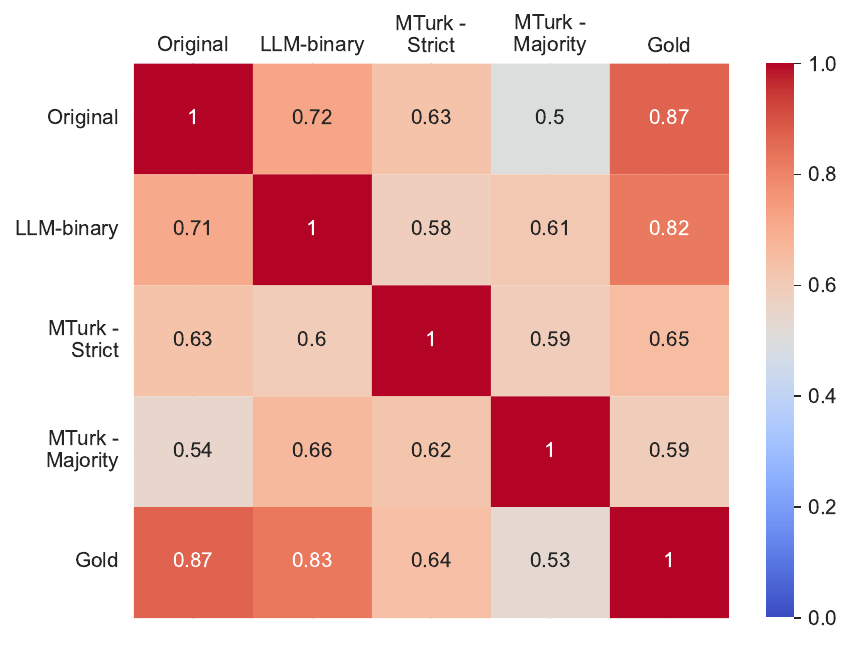}
    \includegraphics[width=0.7\linewidth]{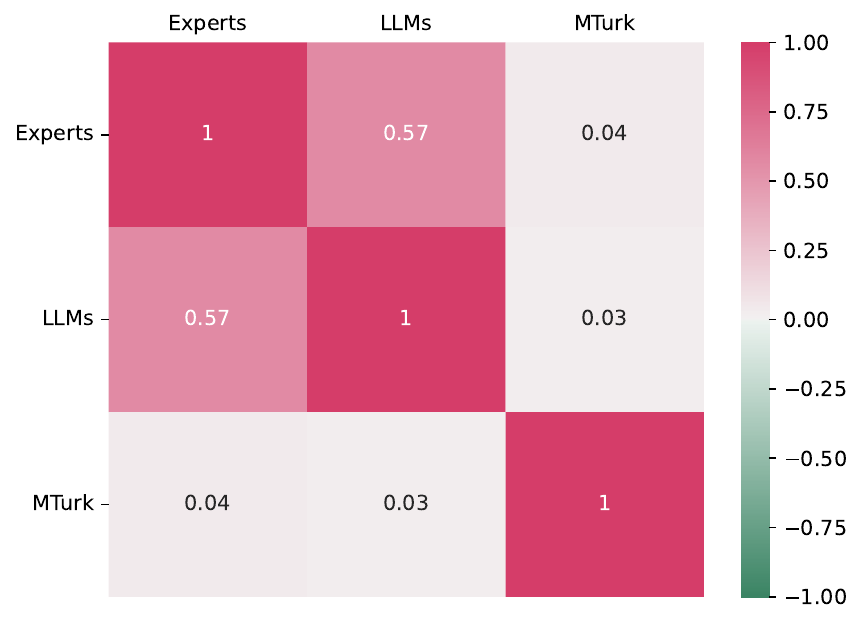}
    \caption{Comparison between all annotation methods: \textbf{(Top)} on the TRUE benchmark, measured by the weighted-F1-score. Rows represent the \textit{"true"} label and columns represent the \textit{"prediction"}. For instance, the score of \emph{LLMs} compared to the \emph{Original} label is 0.72. \textbf{(Bottom)} Comparison on the SummEval benchmark, measured by Pearson correlation (results are averaged over all dimensions).}
    \label{fig:all_comparison}
\end{figure}

\autoref{fig:all_comparison} (top) presents the F1-score between each pair of annotation approaches. As the figure shows, LLMs have disagreements with the \emph{original} labels (0.72). Yet, as discussed in \autoref{sec: label_errors}, the original labels themselves contain mistakes, so this disagreement does not necessarily indicate poor performance of the LLMs. When considering the \emph{Gold} as the true label, LLM performance increases to 0.83. This suggests that LLMs, despite their discrepancies with the original labels, perform closer to the truth than initially reported. The \emph{Gold} label, obtained by experts, has high agreement with both the \emph{Original} and \emph{LLM} labels.
On the other hand, the \emph{MTurk-Majority} approach performs poorly, with near-random F1-scores compared to both the original and gold labels, and even when compared to its stricter variant, \emph{MTurk-Strict}. 
The results indicate that basic crowd-sourcing, without additional training to enhance crowd-workers into specialized sub-experts, performs significantly worse compared to other approaches, including LLM-based methods. On the SummEval dataset (bottom of \autoref{fig:all_comparison}), we observe similar results, where the LLMs are more correlated with the Experts rather than the crowd-workers, which in turn have almost-no-correlation with LLMs or experts' annotations- this implies poor quality of the annotations obtained from crowd-sourcing. \draft{Still, we do not suggest that crowd-sourcing is inherently flawed; with proper task design and worker training, it may be suitable for certain subjective or human-centered tasks. However, we advocate for more careful consideration when using generic crowd annotations for evaluation.}

\paragraph{Crowd-sourcing} For crowd-sourcing, the reported F1-score does not provide the complete picture. When we focus on individual annotators, we see that those who annotate more examples generally deliver higher-quality annotations, achieving greater accuracy when compared to both the original and gold labels (see \autoref{fig:mturk_experience-quality}). This phenomenon can be explained by two hypotheses: (1) a learning process-- as the annotators see more examples, they improve at the task, or (2) users who dedicate time to annotating multiple examples are likely those who either read the guidelines carefully and strive to perform the task to the best of their ability, or are naturally proficient at the task and therefore continue annotating.
Even though annotators who label more instances tend to provide higher-quality annotations, they are less common—most annotators tend to stop after only a few examples. This distribution of annotators results in overall insufficient annotation quality. Pre-qualification tests are often used to shift this distribution from the "average worker" towards more experienced or dedicated annotators; however, this requires a significantly larger budget and greater micro-management involvement from the researcher.

\subsection{Consistency}
\label{sec:consistency}
Usually, when annotating a dataset, more than one annotator is involved. This applies to crowd-workers, experts, and even LLMs- in this study, we use an ensemble of different LLMs and prompts. The use of multiple annotators, similar to an ensemble, is meant to overcome the variance between individuals, which can arise from the subjective nature of NLP tasks, different interpretations of instructions, lack of experience, task difficulty, and cognitive bias \citep{DBLP:journals/jair/UmaFHPPP21}. 

As such, a common practice in the NLP community is to report Inter Annotator Agreement (IAA)- a set of statistical measures used to evaluate the agreement between individuals. Typically, IAA can be viewed as an adjustment of the proportion of pairwise agreements, where 0.0 indicates random agreement. We focus on Fleiss's $\kappa$ \citep{Fleiss1971MeasuringNS}, as it accounts for label imbalance and multiple ($>2$) annotators. High IAA, or low variance between independent annotators, is considered an indicator of high-quality annotation.
In Table \autoref{table: IAA}, we report the agreement between annotators across different approaches. For LLMs, we report two variants: (1) same model, different prompts; and (2) different models, where each model's result is the aggregation across prompts.
For reference, we also include the IAA from the original annotations, as reported in the original papers: \textit{MNBM} reported an average Fleiss's $\kappa$ of 0.696 for the hallucination annotation task; \textit{BEGIN} reported Krippendorff’s $\alpha$ (a generalization of Fleiss's $\kappa$) of 0.7; \textit{VitaminC} reported Fleiss's $\kappa$ of 0.7065 on a sample of 2,000 examples; and \textit{PAWS} reported a 94.7\% agreement between a single annotator’s label and the majority vote on the Wikipedia subset used in TRUE.


\begin{symbolfootnotes}
\begin{table*}[h!]
\centering
\begin{adjustbox}{width=0.6\textwidth}
\begin{tabular}{l|c|c|c}
\toprule
\textbf{Annotator group} & \textbf{Krippendorff's $\alpha$} & \textbf{\%agreement} & \textbf{\#annotators} \\
\midrule
\textbf{Experts} & 0.584 & 60.4 & 3 \\
\midrule
\textbf{MTurk\footnotemark[2]} & 0.496 & 65.6 & 5 \\
\midrule
\textbf{LLM (different prompts)} & & & 4 \\
\quad GPT-4o & 0.760 & 63.6 & \\
\quad Gemini 1.5 Flash & 0.733 & 79.7 & \\
\midrule
\textbf{LLMs (different models)} & 0.576 & 62.9 & 2 \\
\bottomrule
\end{tabular}
\end{adjustbox}
\caption{Inter-Annotator Agreement in different annotator groups on the SummEval benchmark. \%agreement is the proportion of pairwise annotator comparisons.}
\label{table: IAA_summeval}
\end{table*}

\footnotetext[2]{These MTurk annotators were chosen with stricter pre-qualification criteria than those in the TRUE dataset and do not correspond to the MTurk line in the TRUE table.}
\end{symbolfootnotes}

\paragraph{Experts}
While it's true that reconciliation naturally leads to increased agreement, the significant improvement in IAA we observed highlights its importance.
Though this phase is less common in practice, it is crucial not only for increasing agreement but also for improving the overall quality of annotations and ensuring more reliable outcomes.
Interestingly, label changes in this phase were not symmetric, as most changes (69.3\%) were in the direction of \textit{consistent} $\rightarrow$ \textit{inconsistent}, where one annotator found an inconsistency that the other did not (see all change details in  \autoref{fig:experts_changes}). It is important to note that the $\kappa$ obtained by the experts (both before and after reconciliation) was calculated on a more challenging subset, where the original label differed from the LLM prediction, and should be interpreted with this context in mind. This is reflected in the decrease in $\kappa$ observed for all other annotator groups on this subset.

\paragraph{LLMs} 
GPT-4 and PaLM2, the better-performing LLMs on this task, show high IAA, with $\kappa=0.706$ and $\kappa=0.75$, respectively, which is similar to the experts' reported $\kappa$. This suggests a comparable level of variance and quality in annotation, providing further empirical evidence for considering LLMs as annotators. This property adds to previous studies showing LLMs' quality as surrogates for human preferences \citep{llm_as_a_judge} or evaluations \citep{chiang-lee-2023-large}.

\begin{figure}[h]
  \centering
    \includegraphics[width=0.8\linewidth]{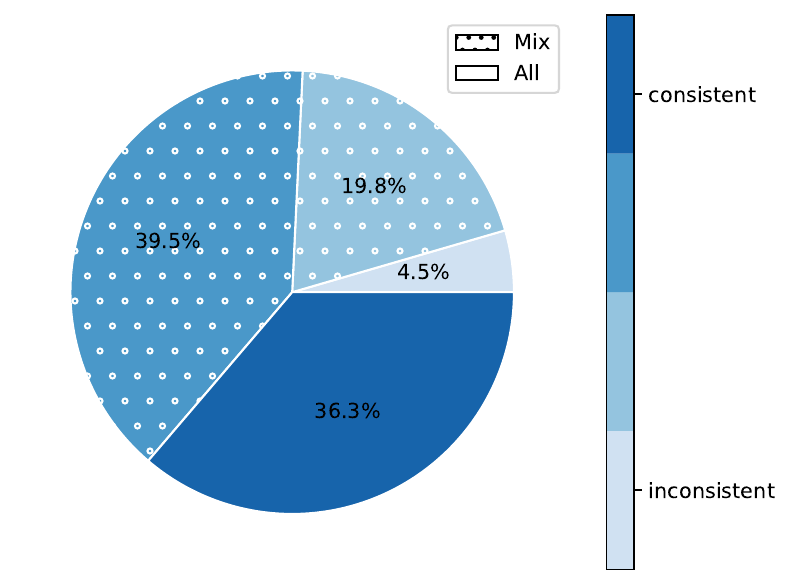}
    \caption{Distribution of crowd-source annotators. Each example was annotated by 3 workers. Plain segments are unanimous annotation, while dotted segments indicate examples where some annotators labeled as \textit{inconsistent}, and other as \textit{consistent}. For example, 19.8\% of the examples had two \textit{inconsistent} annotation, and one \textit{consistent} annotation.}
    \label{fig:mturk_dist}
\end{figure}%

\paragraph{Crowd-Sourcing.} 
Crowd workers showed near-random agreement, indicating relatively poor-quality annotations. \autoref{fig:mturk_dist} describes the distribution of annotations by MTurk workers. Only 40.8\% of the examples were labeled unanimously, whereas the rest included annotations from both classes. In addition, if aggregating by majority vote, we get that 75.8\% of the examples are labeled as \textit{consistent}, which is far from the original distribution of classes. As mentioned before, even experts may miss a small inconsistency nuance, and finding it requires attention. Even from the subset of examples unanimously labeled as \textit{consistent}, 37.9\% have a label of \textit{inconsistent} in both original and gold labels, which points to a lack of attention and thoroughness.
\paragraph{SummEval.}
\autoref{table: IAA_summeval} shows the IAA analysis on the SummEval benchmark. We report Krippendorff’s $\alpha$ \citep{krippendorff1970reliability}, a generalization of $\kappa$ to account for ordinal labeling.
LLMs exhibit high IAA (compared to experts' IAA) of $\alpha=0.57$ and 62.9\% agreement between models, with high consistency across prompts for the same model. Crowd-workers obtain decent results (maybe due to stricter pre-qualification criteria of 10,000 approved HITs), yet they still fall short compared to experts or LLMs.

\subsection{Cost and Scalability}
\label{sec: cost-scale}
In MTurk platform, a total of $400\times3=1200$ annotations cost $572 \$$, including 2 small pilot experiments. All annotations were prepared within a few hours. However, it demanded an additional and significant time for review, after which rejected examples returned to the pool. This annotation-review cycle was conducted for $\sim 5$ iterations.
Inference via OpenAI's API on GPT-4 cost $\sim 4.5 \$$ per prompt.
Inference via VertexAI's API on \codefont{PaLM2} cost $\sim 0.15 \$$ per prompt. Both took $\sim 8$ minutes per prompt.
Inference on \codefont{Mistral} and \codefont{Llama3} was via the HuggingFace API, and its cost is estimated by the cost of using a suitable Virtual Machine (VM) on Google Cloud Platform (GCP) for the time of inference (1 minute per model)- $\sim 0.1 \$$ per prompt.

LLM-based annotation is significantly cheaper and faster than crowd-sourcing platforms like MTurk, especially when considering the additional time required for human review cycles. It is estimated to be 100 to 1,000 times more cost-effective than using human annotators, including experts. This scalability and speed make LLMs a highly efficient alternative for large-scale annotation tasks.

\section{Annotation}
\label{sec: appendix-annotation}
\subsection{Crowd-source}
\label{appendix-crowd-source}
\begin{figure*}[t]
    \centering
    \includegraphics[height=0.45\textheight]{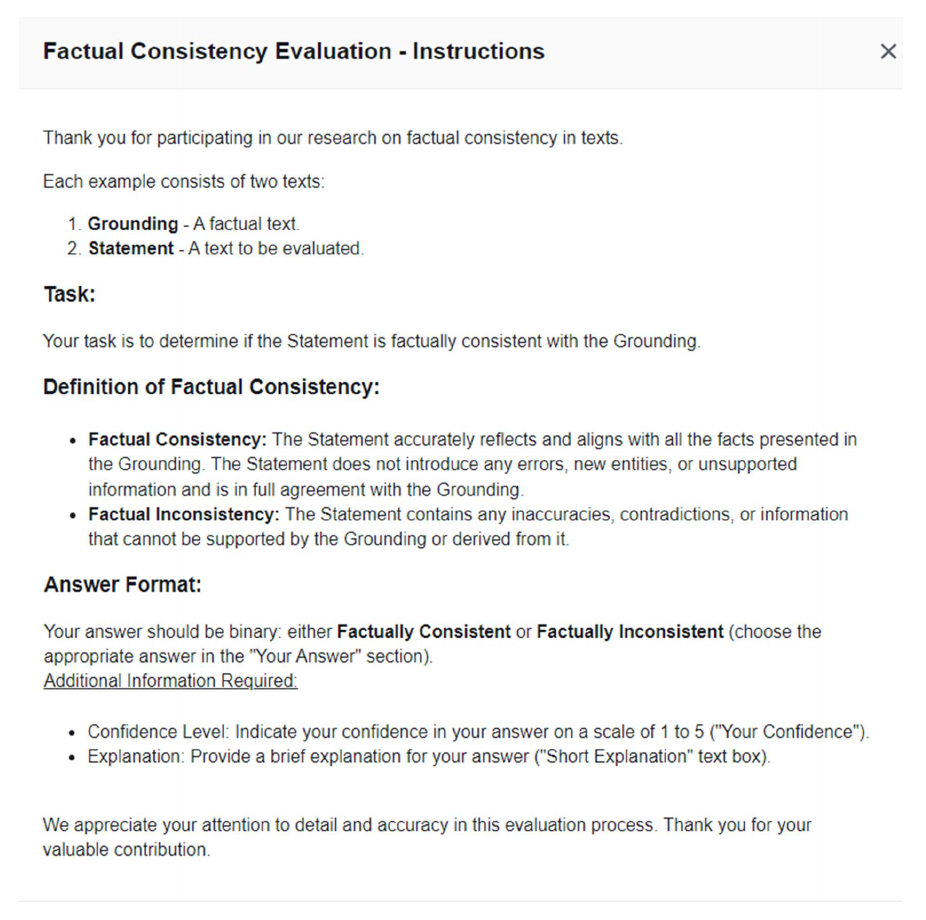}
    \includegraphics[width=0.8\textwidth]{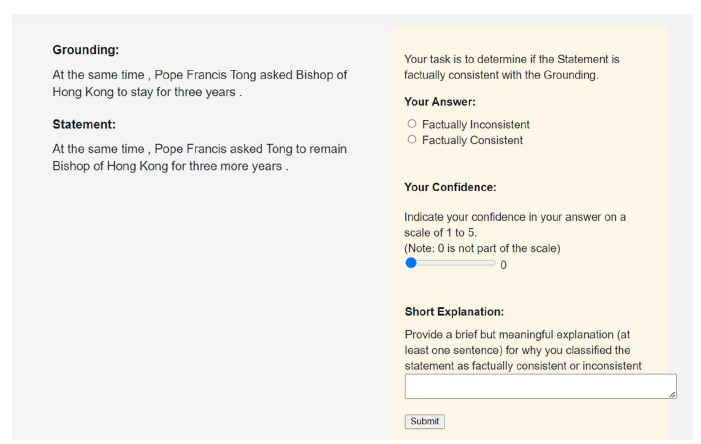}
    \caption{Platform for crowd-sourcing annotation in Amazon Mechanical Turk (MTurk). \textbf{(Top)} Guidelines for the task and definitions. \textbf{(Bottom)} Annotation layout for a single instance.}
    \label{fig:mturk_platform}
    \vspace{-2em}
\end{figure*}

Each example was annotated by three annotators, who in addition to the binary label were requested to provide their confidence in their answer, and also write a short explanation for why they chose this label.
Pre-qualifications included 50+ approved HITs and 97\%+ approval rate, which are at standard scale for the MTurk platform \citep{DBLP:journals/ir/KazaiKM13, Hauser2021EvaluatingCA, Chmielewski2019AnMC}. Also, locations were limited to [USA, UK, Australia], which are all English-speaker countries.
We disabled the possibility of right-click and \codefont{Ctrl+c} in the platform (as suggested by \cite{veselovsky2023prevalencepreventionlargelanguage}), to prevent (as much as possible) the case where generative-AI (e.g., ChatGPT) will be applied to solve the task instead of humans solving it themselves (as shown by \citep{artificial_users}).
The maximum time allowed per HIT was 6 minutes, while the actual average execution time was 2:20 minutes for all assignments, and 3 minutes for approved assignments.
The guidelines provided to annotators and the annotation platform layout are presented in \autoref{fig:mturk_platform}.

Each annotation was manually reviewed and was rejected if the answers were not in line with the instructions, or if it was obvious that the task was not done honestly. Overall, this task suffered from a high rejection rate of 49.2\% (1163 rejected, 1200 approved). The main rejection reasons were:
lack of meaningful explanation, obvious copy-paste annotations across different examples, explanations contradicting the label annotation, and cases where the explanation was a copy-paste of either the grounding or the statement.

\clearpage
\clearpage

\twocolumn[{
    \begin{center}
    \includegraphics[width=0.8\linewidth]{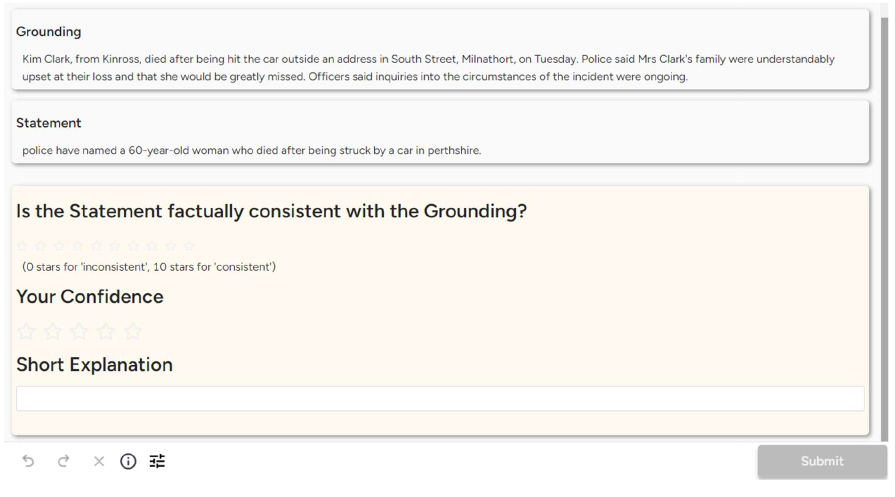}
    \end{center}
    \captionof{figure}{Annotation platform on Label-Studio for experts}
    \label{fig:experts_platform}
    \vspace{1em}
    
}]

\subsection{Experts}
\label{appendix-experts}
Experts annotation was using the platform of Label Studio. \footnote{\url{https://labelstud.io/}} Layout design is presented in \autoref{fig:experts_platform}. Examples were presented in random order, and neither the LLM prediction nor the original label were presented during the annotation. In the first stage, each example was annotated independently by both experts. Afterward, the human experts began in a second phase of a reconciliation, where a discussion was made over examples they disagreed over. This reconciliation phase ended up with a much higher agreement and higher-quality labels. 
\draft{Complete agreement was reached in nearly all cases; only a very small number of examples remained unresolved, which may reflect inherent label variation rather than clear annotation errors \citep{weber-genzel-etal-2024-varierr}.}

\begin{figure}[h]
\vspace{-1em}
  \centering
  \includegraphics[width=0.7\linewidth]{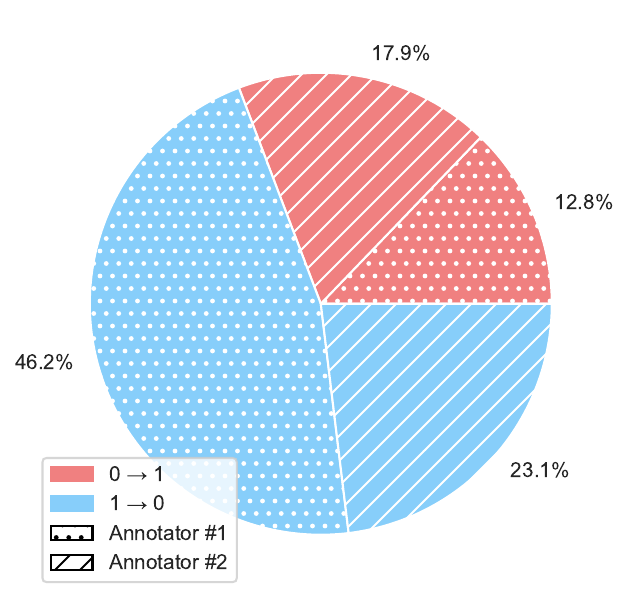}
    \caption{How experts' annotations have changed after the reconciliation phase. Most changes occur from 1 (\textit{consistent}) to 0 (\textit{inconsistent}).}
    \label{fig:experts_changes}
    \vspace{-1em}
\end{figure}

In the reconciliation phase, we observed that most changes (69.3\%) were from label 1 to label 0, indicating that contradictions might be hard to find, and not all annotators catch them at first. For the full distribution of label change in the reconciliation phase, see \autoref{fig:experts_changes}.

\subsection{LLMs}
\label{sec: appendix-llms}

To annotate a total of $160\times4=640$ examples from four different datasets, we used four LLMs: GPT-4 (\codefont{gpt-4-1106-preview}) \citep{gpt4}, PaLM2 (\codefont{text-bison@002}) \citep{palm2}, Mistral (7B)\footnote{\url{https://huggingface.co/mistralai/Mistral-7B-Instruct-v0.2}} \citep{mistral7b} and Llama 3 (8B)\footnote{\url{https://huggingface.co/meta-llama/Meta-Llama-3-8B-Instruct}} \citep{llama3}.

\begin{figure*}
    \centering
    \includegraphics[width=0.9\textwidth]{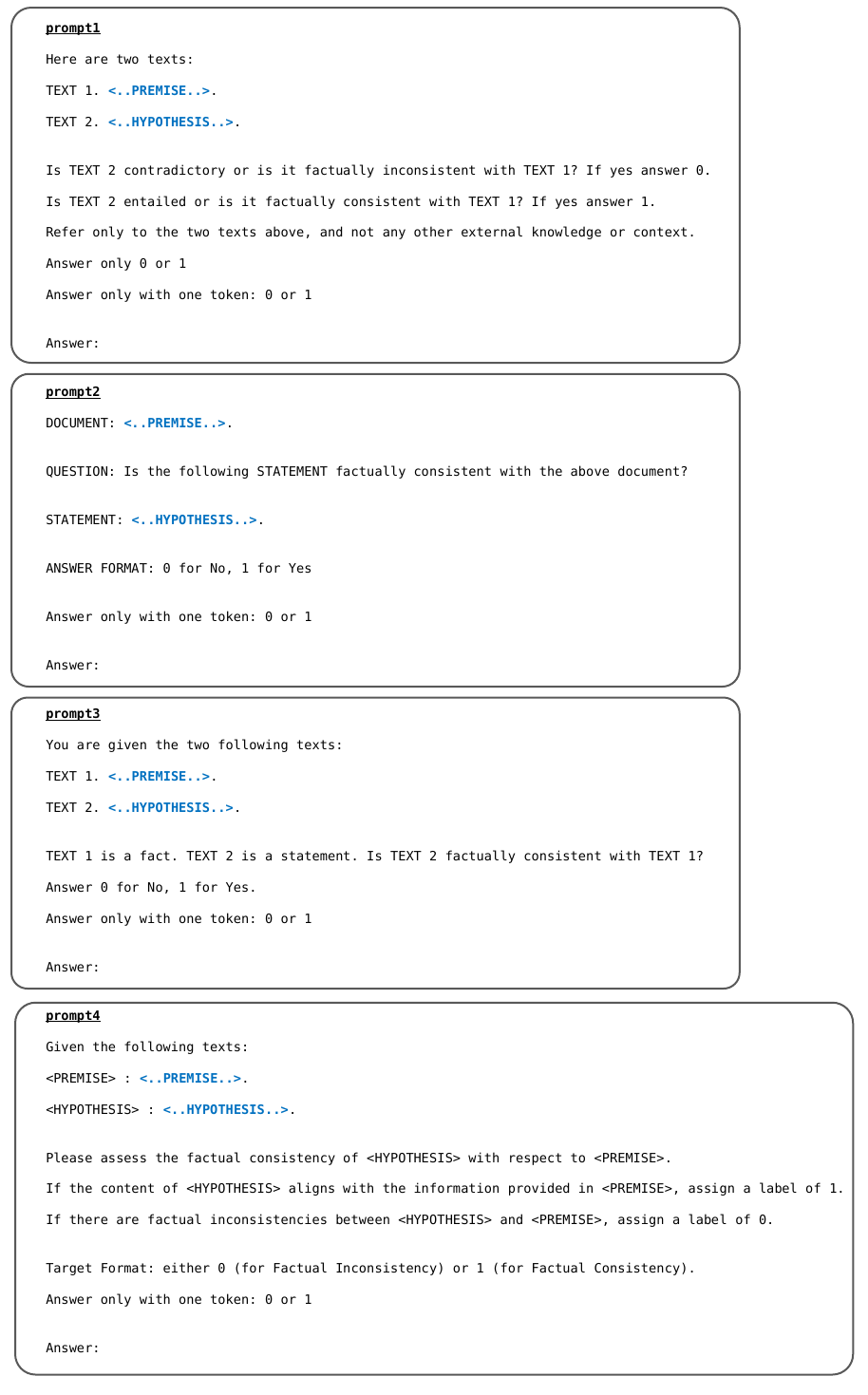}
    \caption{Four different prompt input templates to LLMs for obtaining binary labels}
    \label{fig:prompts}
\end{figure*}

Each model was run with four different prompts (see full prompts in \autoref{fig:prompts}). We used a variety of terminology, as this task appears to have different framings in different studies. For example, the premise-hypothesis terminology from classic NLI \citep{maccartney-manning-2009-extended}, or document-statement used in \citep{tam-etal-2023-evaluating}. The ensemble reported in the main text refers to ensembling GPT-4 and PaLM2 over four prompts, while the other models are intended for extending our analysis to more models.

For API models (GPT-4, PaLM2), we set \codefont{temperature=0.0} and extracted the logit of the generated token (functionality provided by both APIs), if the generated token was either \codefont{'0'} or \codefont{'1'} as expected. This logit was then transformed into a probability $p_t = P(y=t|x)$ via exponent corresponding the generated token $t$, and $1-p_t$ for the other label.
To address the case where the first generated token was an unrelated token such as \codefont{' ', '\textbackslash n'}, we set \codefont{max\_tokens=2} and took the first appearance of either \codefont{'0'} or \codefont{'1'}. For all models, prompts and examples, \codefont{'0'} or \codefont{'1'} were in the first two generated tokens. Rest of parameters were set according to their default values.

For models available through the HuggingFace API (e.g., Mistral, Llama 3), we can load the model parameters and make inference locally. In that case, we get access to logits for all tokens, instead of just for the generated ones. Therefore, we applied a similar procedure, where we seek for the first appearance of either \codefont{'0'} or \codefont{'1'} to be the most probable token to be generated, and then directly extracted the logits of the \codefont{'0'} and \codefont{'1'} tokens. These logits were transformed into probabilities $(P(y=0|x), P(y=1|x))$ via a softmax function.

\section{Data}
\label{sec: appendix_data}
For our main experiments, we used the TRUE benchmark for factual consistency.
Specifically, we focus on four TRUE datasets, one from each task (summarization, dialogue, fact verification, paraphrasing):

\paragraph{MNBM \citep{MNBM}: Summarization.}
This dataset provides annotations for hallucinations in generated summaries from the XSum dataset \citep{xsum}. \emph{Grounding} refers to the source document that the summary is based on, while \emph{Generated Text} consists of model-generated summaries, which may include hallucinated information not present in the source. Three human annotators, trained for the task through two pilot studies, annotated the dataset for the existence of hallucinations. In TRUE, the binary annotations were determined by majority vote.
    
\paragraph{BEGIN \citep{BEGIN}: Dialogue.}
This dataset evaluates groundedness in knowledge-grounded dialogue systems, where responses are expected to align with an external \emph{Grounding} source, typically a span from Wikipedia. \emph{Generated Text} refers to model-generated dialogue responses that were fine-tuned on datasets like Wizard of Wikipedia \citep{WoW}.
Data was annotated into entailment/neutral/contradiction labels, by three human annotators, trained for the task through two pilot studies, aggregated by majority vote. In TRUE, binary annotations were then determined by the entailment/not-entailment partition.
    
\paragraph{VitaminC \citep{vitc}: Fact Verification.} 
This dataset is based on factual revisions of Wikipedia. The evidence, or \emph{Grounding}, consists of Wikipedia sentences, either before or after these revisions. Most human involvement came from creating \emph{Generated Text} rather than the annotation process, with annotators writing claim/evidence pairs derived from Wikipedia revisions, inherently generating labeled data for fact verification. Synthetic examples from the FEVER dataset \citep{thorne-etal-2018-fever} were also included. Additionally, three annotators reviewed 2,000 examples, presumably to ensure data quality.
    
\paragraph{PAWS \citep{PAWS}: Paraphrasing.} 
This dataset consists of paraphrase and non-paraphrase pairs. \emph{Grounding} refers to source sentences drawn from Quora and Wikipedia, while \emph{Generated Text} was automatically generated through controlled word swapping and back-translation.
Five human annotators annotated the dataset with binary labels w.r.t paraphrasing correctness. The dataset includes both high- and low-agreement annotations.

\clearpage
\twocolumn[{
  \centering
    \includegraphics[width=0.48\textwidth]{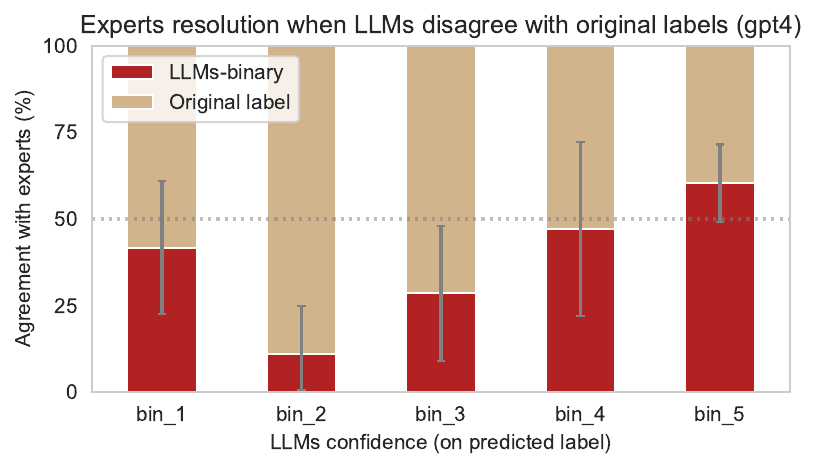}
    \includegraphics[width=0.48\textwidth]{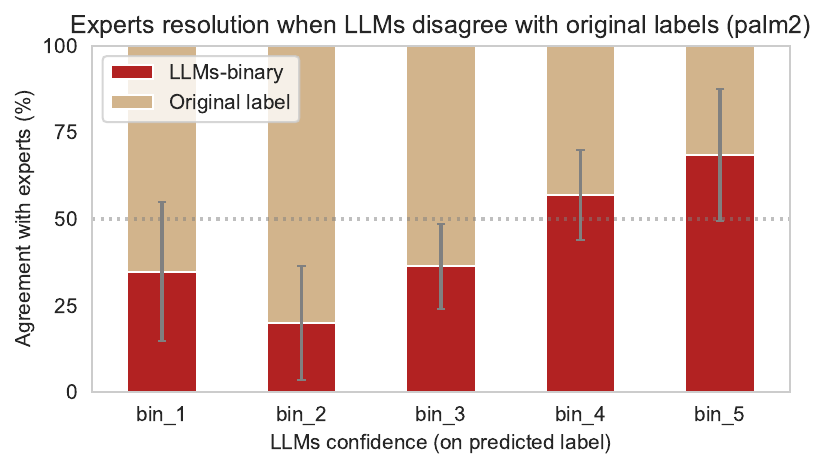}
    \includegraphics[width=0.48\textwidth]{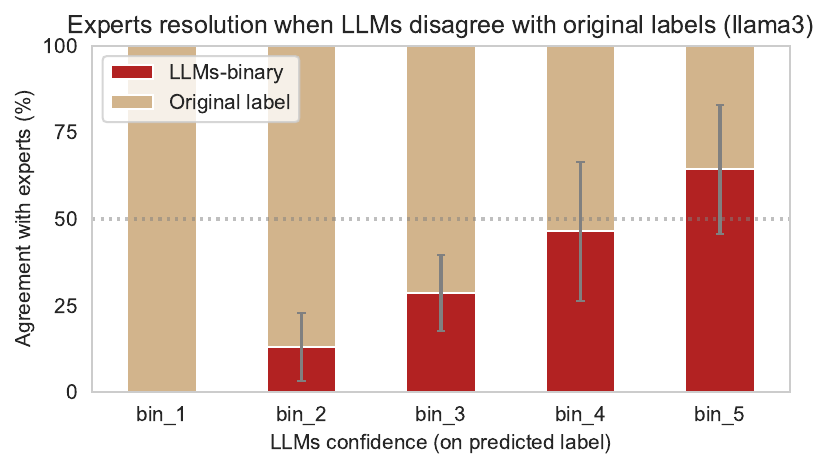}
    \includegraphics[width=0.48\textwidth]{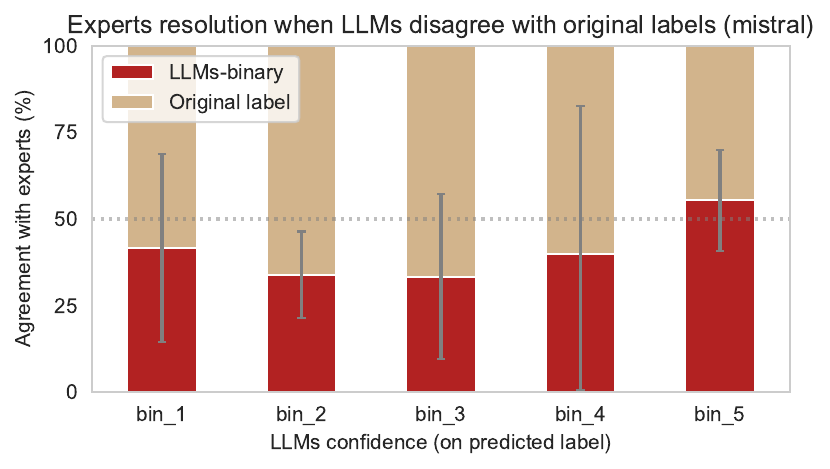}
  \captionof{figure}{A model-specific analysis: When LLMs disagree with original labels - who is correct? Overall trend holds: as the LLM's confidence grows, so does the precision of identifying an error in the original labels. To maintain a balanced number of examples per bin, the bin edges from Figure~\ref{fig:experts_resolution} were slightly adjusted per model due to natural variability and calibration differences. For simplicity in the shared plot across all models, we label the bins as bin1 through bin5, where bin1 starts at 0.5 and bin5 ends at 1.0.}
  \label{fig:appendix_expert_resolution_specific}
\vspace{2em}
}]

\section{Model-Specific Experiments}
\label{sec:appendix_model_specific}

Our main analysis relies on an ensemble-based approach, which abstracts away from individual model behavior and leverages their collective strength. This design improves alignment with expert annotations, reduces variance, and avoids the need for model selection or prompt-specific tuning. As such, it provides a more stable and generalizable signal than any single model. The ensemble results are presented in \autoref{fig:experts_resolution}.

For completeness, we also provide a model-specific analysis of the same phenomenon. \autoref{fig:appendix_expert_resolution_specific} reports the percentage of cases where experts agreed with the LLM prediction rather than the original label, broken down by confidence bins and shown separately for GPT-4, PaLM2, LLaMA-3, and Mistral. These curves correspond directly to the red bars in \autoref{fig:experts_resolution}, but now reveal each model’s contribution.

Across models, we observe the same overall trend: when models express higher confidence in a label that differs from the original annotation, experts are increasingly likely to agree with them. The magnitude and variance of this effect, however, differ by model. Some models, such as LLaMA-3, display clearer calibration, while others, such as Mistral, show flatter patterns.

A single model is cheaper and can capture the main trend, but its behavior varies by model. Our ensemble reduces these differences, yielding smoother calibration (\autoref{fig:experts_resolution}), more consistent agreement with experts, and lower variance—benefits we believe justify the extra compute.

\twocolumn[{
    \centering
    \includegraphics[width=0.45\linewidth]{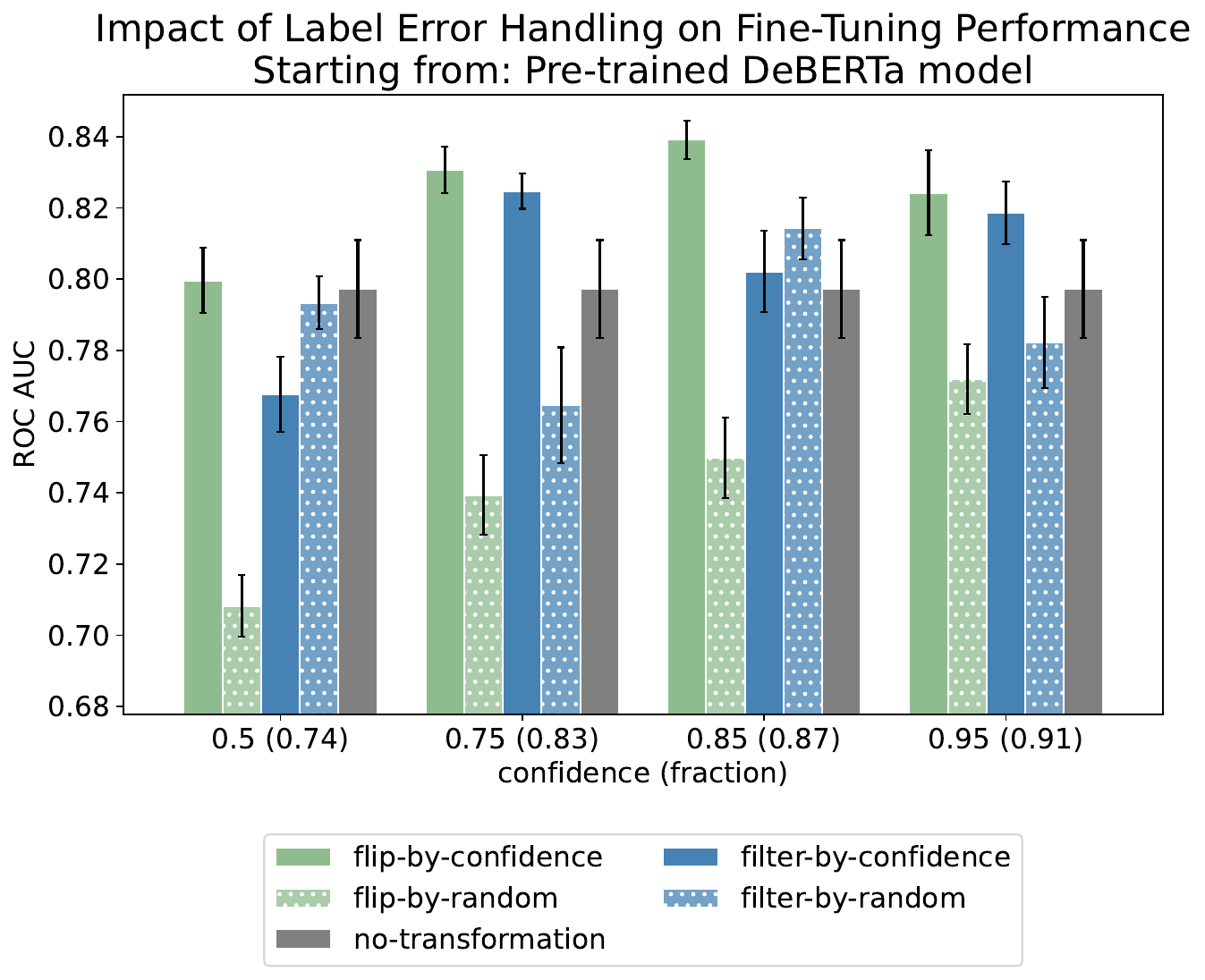}
    \includegraphics[width=0.45\linewidth]{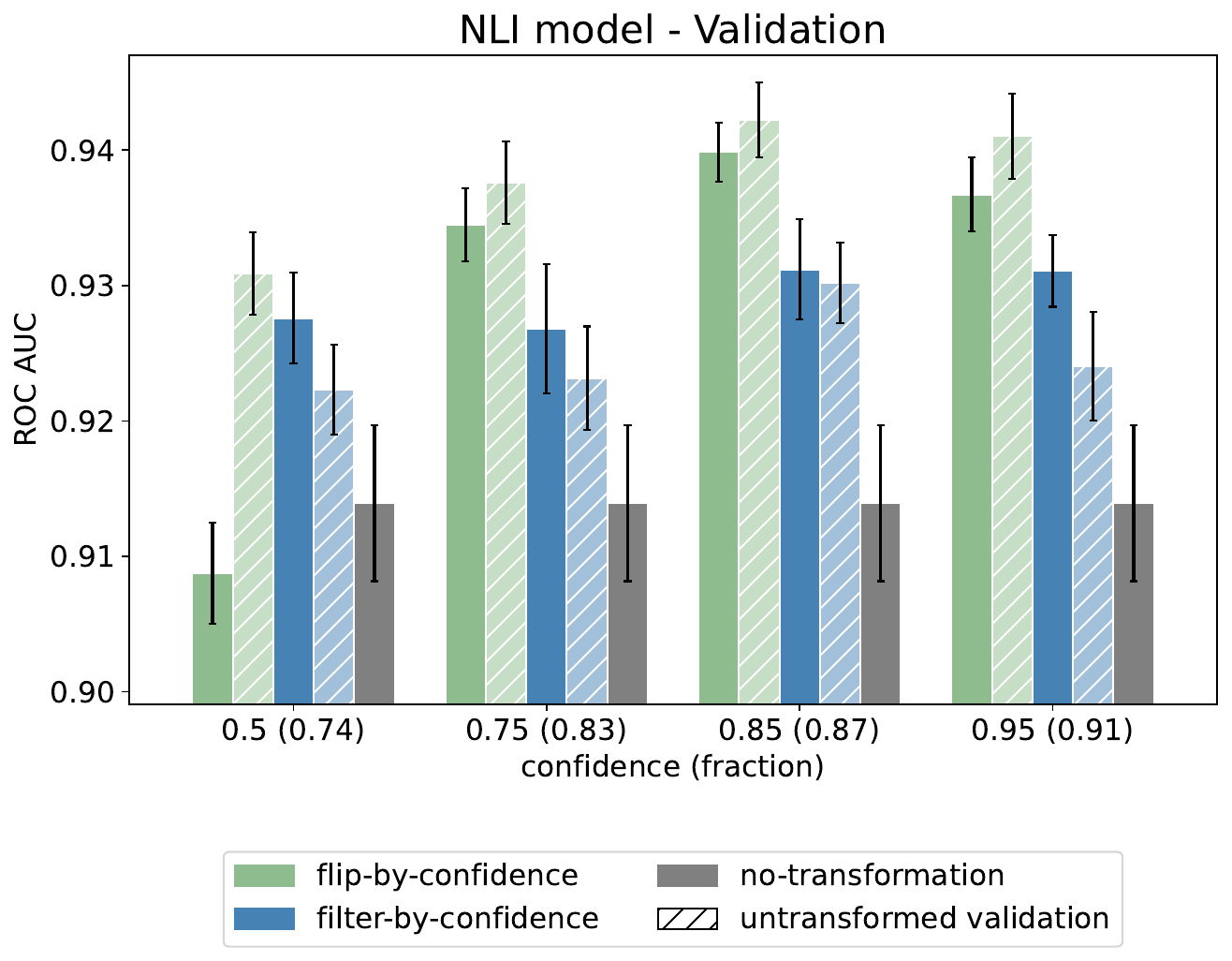}
    \captionof{figure}{Similar experiments to the one in \autoref{fig:finetuning}, with small alterations. \textbf{(Left)} Starting from a different base model - pre-trained \codefont{DeBERTa-v3-base}. \textbf{(Right)} Dashed columns present results for when flipping or filtering methods were applied only on the training set, but not the validation.}
    \label{fig:finetuning_additional}
    \vspace{2em}
}]

\section{Mislabeled Data Implications}
\subsection{Fine-tuning}
\label{appendix-finetuning}
\paragraph{Hardware.} For the finetuning of DeBERTa models, both the base pre-trained model, and the NLI model which is in the same size, in \autoref{sec: implications-training}, we used 2 Quadro RTX6000 (24GB) GPUs.

\paragraph{Implementation.} We finetuned starting from two base models: DeBERTa-v3 \footnote{\href{https://huggingface.co/microsoft/deberta-v3-base}{\texttt{microsoft/deberta-v3-base}}}, and a fine-tuned version of it on classic NLI datasets \footnote{\href{https://huggingface.co/MoritzLaurer/DeBERTa-v3-base-mnli-fever-anli}{\texttt{MoritzLaurer/DeBERTa-v3-base-mnli-fever-anli}}}. We used HuggingFace trainer with early stopping of 4 epochs. The finetuning procedure includes splitting the training set into train and validation sets (where validation size is 25\% and train 75\%), fine-tuning on the train set, and choosing the best checkpoint based on the validation ROC AUC. We ran all experiments on five different seeds, affecting also the train-validation split and the random set chosen for ablation. We fine-tuned all variants with the same hyperparameters, determined by the best performing on the no-manipulation baseline. This includes 30 epochs at most, batch size of 16, learning rate of \codefont{5e-5} and weight-decay of 0.03. The rest were set as the trainer and model default.

\paragraph{Additional Experiments.}
The left plot in \autoref{fig:finetuning_additional}  presents the same experiment discussed in \autoref{sec: implications-training}, but starting from the pre-trained \codefont{DeBERTa-v3-base}. Same trends applies here, where our LLM-confidence-based manipulations of either flipping or filtering flagged examples outperforms the baselines.

The right plot in \autoref{fig:finetuning_additional} compares the performance of these methods (starting from the NLI model) when applied to both the training and validation sets (solid bars) or only the training set (dashed bars). The results are consistent, with no statistically significant differences between the two settings. Importantly, all variations outperform the baseline, underscoring the critical role of a well-curated training set in enhancing the model's ability to generalize effectively.

\subsection{Model Evaluation}
\label{sec: appendix-eval}
In \autoref{sec: implications-eval} we evaluated the following models: GPT-4, PaLM2 (\codefont{text-bison@002}), Mistral-v0.2 (7B), and Llama3 (8B), which are covered in \autoref{sec: annotation-llm}; DeBERTa-v3 and NLI-model, which is a fine-tuned version of it on NLI datasets, as discussed in \autoref{sec: implications-training}; and GPT-4o, GPT-4o-mini, Mistral-v0.3,\footnote{\url{https://huggingface.co/mistralai/Mistral-7B-Instruct-v0.3}} which share the same implementation as GPT-4 or Mistral-v0.2.

\draft{
\paragraph{Fine-Tuning vs. Zero-Shot}
Interestingly, the overall trend of improved performance on the corrected labels does not hold for the DeBERTa-based fine-tuned models. Unlike the LLMs, which are prompted in a zero-shot setting, the fine-tuned models are trained on the original dataset, which contains label errors. As a result, the LLMs demonstrate better generalization, while the fine-tuned models may overfit to the noise in the training data. A plausible explanation for this reversed trend lies in the distributional prior learned from the training set. In the original dataset, labels of 0 (inconsistent) are more frequent than in the corrected gold set. For example, among examples where the original and gold labels agree, the proportion of 1 (\textit{consistent}) labels is 36\%, and the model (\codefont{DeBERTa-v3-base} predicts 1 in 35\% of those cases. In contrast, among examples where the labels disagree, the gold rate of 1 is 58\%, yet the model predicts 1 in only 36\% of the cases. This pattern suggests that the model has learned a skewed prior from the flawed dataset, underestimating the likelihood of the consistent class, particularly in cases that were originally mislabeled. Similar percentages are observed for the NLI model as well.
}

\section{Statistical Analysis}
\label{sec: appendix-stats}
\subsection{Clopper-Pearson}
\label{sec: appendix-CI_error_rate}
As mentioned in \autoref{sec: label_errors}, we employed the Clopper-Pearson exact method \citep{clopper-pearson} to construct a 95\% confidence interval for the binomial proportion, adjusted by a finite population correction (FPC). As we only have a subset of examples we re-annotated by LLMs or experts, we can not precisely determine what is the error rate in the full dataset, but only construct a confidence interval based on the re-annotated subset. 
The Clopper-Pearson method provides an exact confidence interval for a binomial proportion, which means it gives a reliable estimate even with small sample sizes. By applying FPC, we adjust the interval because our sample is drawn from a limited population. This adjustment helps refine the estimate by taking into account the size of the overall dataset compared to the sample.




\subsection{Bootstrap sampling}
\label{sec: appendix-bootstrap}
In \autoref{sec: label_errors}, we use bootstrap sampling to provide confidence intervals for each bin. While not necessarily the first to introduce it, \cite{Xia2012TranslationalBD} explored bootstrap confidence intervals on ROC AUC. Unlike the method in Appendix \ref{sec: appendix-CI_error_rate}, we do not make claims about the entire dataset, but rather focus on the re-annotated subset we possess. To achieve this, we perform 100 bootstrap samples from the empirical distribution of each bin, sampling with replacement. We then measure the agreement between the experts' resolutions and the LLM annotations, compared to its agreement with the original label.

\section{Label Errors}
\autoref{table: annotation_errors_full} demonstrates one example per dataset, in which the original label is, in fact, an error, the LLM prediction marked it as a candidate, and the expert annotators determined the correct gold label.
\begin{table*}[h]

\centering
\renewcommand{\arraystretch}{1.2}
\begin{adjustbox}{max width=\textwidth}
\begin{tabular}{|p{15cm}|}
        \hline
    \textbf{Dataset:} VITC \\
    \textbf{Grounding:} \small The British Government and NHS have set up a Coronavirus isolation facility at Arrowe Park Hospital in The Wirral for British People coming back on a special flight from Wuhan. Evacuation of foreign diplomats and citizens from Wuhan. Due to the effective lockdown of public transport in Wuhan and Hubei province , several countries have started to evacuate their citizens and/or diplomatic staff from the area , primarily through chartered flights of the home nation that have been provided clearance by Chinese authorities. \\
    \textbf{Generated Text:} \small There is a Coronavirus isolation facility at Arrowe Park Hospital that was set up by the NHS and the British Government \\
    \textbf{Original Label:} 0 \hspace{1cm} \textbf{LLM $p$:} 0.99  \hspace{1cm} \textbf{Gold Label:} 1 \newline \greyfont{\textbf{Explanation}: Rephrasing of the first sentence, without any contradiction.} \\
    \hline
    \textbf{Dataset:} BEGIN \\
    \textbf{Grounding:} \small Hillary Clinton, the nominee of the Democratic Party for president of the United States in 2016, has taken positions on political issues while serving as First Lady of Arkansas (1979–81; 1983–92), First Lady of the United States (1993–2001);  \\
    \textbf{Generated Text:} \small She is the nominee in 2016. \\
    \makecell[l]{\textbf{Original Label:} 0 \hspace{1cm} \textbf{LLM $p$:} 0.98 \hspace{1cm} \textbf{Gold Label:} 1} \newline
    \greyfont{\textbf{Explanation}: She (Hillary Clinton) is indeed the nominee in 2016 as specifically stated in the grounding.} \\
    \hline
    \textbf{Dataset:} PAWS \\
    \textbf{Grounding:} \small David was born in Coventry on 21 September 1933 , with his twin Charles and Jessamine Robbins , the eighth and ninth children of twelve by Robbins. \\
    \textbf{Generated Text:} \small David was born on September 21 , 1933 in Coventry with his twin father Charles and Jessamine Robbins , the eighth and ninth child of twelve of Robbins \\
    \textbf{Original Label:} 1 \hspace{1cm} \textbf{LLM $p$:} 0.04 \hspace{1cm} \textbf{Gold Label:} 0 \newline
    \greyfont{\textbf{Explanation}: The generated text incorrectly states "twin father" instead of "twin" which is not the same, and does not even make much sense in English.} \\
    \hline
    \textbf{Dataset:} MNBM \\
    \textbf{Grounding:} \small The John Deere tractor was pulled over by officers in the village of Ripley and had two other males on board. The vehicle had been seen in nearby Harrogate at about 05:00 GMT with no headlights on. Police said the driver had no licence, was not insured and did not have permission from the tractor's owner. The vehicle was seized, with the three due to be interviewed by officers. Posting on Twitter, Insp Chris Galley said: "A strange end to a night shift. 15-year-old lad driving a tractor as a taxi for his drunk mates." \\
    \textbf{Generated Text:} \small a 15-year-old boy has been stopped by police after being seen driving a taxi on a night taxi. \\
    \textbf{Original Label:} 1 \hspace{1cm} \textbf{LLM $p$:} 0.19 \hspace{1cm} \textbf{Gold Label:} 0 \newline
    \greyfont{\textbf{Explanation}: The generated text claims that the 15-year-old boy was "driving a taxi on a night taxi", contradicting the grounding in which it was claimed that the boy was driving a tractor as a taxi} \\
    \hline

\end{tabular}
\end{adjustbox}
\caption{Annotation errors in the original datasets, discovered by LLMs and corrected by experts.}
\label{table: annotation_errors_full}
\end{table*}
\end{document}